\documentclass[11pt]{article}
\usepackage[preprint]{acl}
\usepackage{times}
\usepackage{latexsym}
\usepackage[T1]{fontenc}
\usepackage[utf8]{inputenc}
\usepackage{microtype}
\usepackage{graphicx}
\usepackage{booktabs}
\usepackage{multirow}
\usepackage{float}
\usepackage{amsmath}
\usepackage{enumitem}
\usepackage{amssymb}
\usepackage{algorithm}
\usepackage{algpseudocode}
\usepackage{xcolor}
\usepackage{colortbl} 
\usepackage{tikz}
\usepackage{pgfplots}\pgfplotsset{compat=1.17}
\usepackage{pifont}
\definecolor{gkNavy}{HTML}{1D437D}\definecolor{gkNavyBg}{HTML}{DDE3EC}
\definecolor{gkGreen}{HTML}{0D7B34}\definecolor{gkGreenBg}{HTML}{97C8AF}
\definecolor{gkRed}{HTML}{EE442F}\definecolor{gkRedBg}{HTML}{FDE3E0}
\definecolor{gkGray}{HTML}{808080}\definecolor{gkOrange}{HTML}{FFA500}
\definecolor{gkTeal}{HTML}{0099AD}\definecolor{gkYellow}{HTML}{DEC923}
\definecolor{cbBlue}{HTML}{4C72B0}\definecolor{cbBlueBg}{HTML}{A6B9D8}
\definecolor{fig1Blue}{HTML}{99CCFF}
\definecolor{cbRed}{HTML}{C44E52}\definecolor{cbRedBg}{HTML}{E2A7A9}
\definecolor{cbGreen}{HTML}{55A868}\definecolor{cbGreenBg}{HTML}{AAD4B4}
\usetikzlibrary{calc}
\usetikzlibrary{patterns}
\usetikzlibrary{decorations.pathreplacing}
\usetikzlibrary{positioning,arrows.meta,fit,calc,decorations.pathreplacing,backgrounds}

\newcommand{\pending}[1]{}
\newcommand{\ours}{Oilbird}

\title{Oilbird: Training-Free Speculative Decoding \\ with Keys the Verifier Already Computes}

\author{
  Tao Jin\textsuperscript{1} \quad Phuong Minh Nguyen\textsuperscript{1} \quad Zhenzhu Yan\textsuperscript{1}
  \quad Teeradaj Racharak\textsuperscript{2} \quad Naoya Inoue\textsuperscript{1}\thanks{\ Corresponding author.} \\[2pt]
  \textsuperscript{1}Japan Advanced Institute of Science and Technology (JAIST) \quad
  \textsuperscript{2}Tohoku University \\[2pt]
  \texttt{\{morgan, phuongnm, s2420001, naoya-i\}@jaist.ac.jp} \quad
  \texttt{racharak@tohoku.ac.jp}
}

\begin{document}
\raggedbottom 
\maketitle

\begin{abstract}
Training-free speculative decoding drafts by matching an exact suffix of the context against a pool
of earlier context. That lookup misses correct drafts already in the pool, most visibly on
tool-calling traffic, where a request repeats almost everything but the few values minted for it,
and where one rejected token discards the correct continuation behind it. We diagnose the failure
position by position across ten benchmarks and find it to be a problem of addressing rather than of
coverage: on our densest tool-calling benchmark, about half of what the strongest exact-match
drafter misses is present in the pool yet unreachable by exact matching. We therefore propose a second,
semantic draft source: the same pool, re-keyed by the hidden state the verifier has already
computed at each committed token, together with a merge that lets it ride inside an existing lexical
drafter's tree. In three published drafters, at matched pool and budget, it lifts accepted length by
24--29\%. \ours{} reaches 4.4$\times$ autoregressive decoding speed on API-Bank, against 3.9$\times$ for the strongest training-free baseline in our harness and
2.0$\times$ for EAGLE-3.
\end{abstract}

\section{Introduction}
\label{sec:intro}

\emph{Speculative decoding} (SD) uses a cheap \emph{drafter} to propose future tokens that the target verifies in one forward pass, committing the matching prefix: under greedy verification the output is unchanged (\emph{lossless}) \citep{leviathan2023fast}, and under sampling a modified rejection rule preserves the target's distribution \citep{chen2023accelerating}. Modern implementations verify a tree of drafts per step \citep{miao2024specinfer,chen2024sequoia}. The central quantity is the \emph{accepted length} $\tau$, the tokens committed per verification pass.
Drafters are either trained or training-free; the training-free ones copy earlier text, so they are strongest where serving traffic, the stream of requests a deployed model answers, is most redundant: code, math, and above all tool-calling agents, which emit near-identical API calls at volume and are waited on by a user, where a lossless drafter buys back time without changing any output.

Where requests repeat, an exact-suffix key matches almost everywhere, nearly all of a call being text the service has emitted before, and fails chiefly at the few tokens minted for this request: an identifier, a name. Those are in no past trace, and so beyond any drafter that \emph{copies}, however perfect its retrieval. Nor does the key degrade there: it dies. One such token drives every suffix containing it to zero matches, at every length. Call such a position
\emph{blind}: what it costs is not the token it misses. Behind it the drafter already held the frame that resumes once the value has passed, and committing only the matching prefix discards that frame too.

A miss is then one of two things: an absent continuation, or a present one the key cannot name. The standing answer grows the store \citep{he2024rest,oliaro2024suffixdecoding}; we separate the two, position by position (Figure~\ref{fig:gapproblem}). On API-Bank, our densest tool-calling benchmark, about half of what the strongest exact-match drafter we can build misses is present but unreachable by exact matching: the \emph{identifiability gap}. A key that matches on the model's own state instead of the text, under the same candidate budget, reaches the true next token at four in five of those positions, behind which a neighbour matches 6.4 more tokens verbatim. The constraint there is addressing, not coverage.

\ours{} tags every committed token with the hidden state the verifier has already computed and retrieves past positions whose states match. We then design the merge that spends those continuations: they re-walk a multi-source attention tree from the root beside the lexical sources already in it, paying for a node only where they diverge, so the source rides at an unchanged node budget with verification cost and the acceptance rule untouched. The output is audited token for token against an independent greedy run.
Two choices are load-bearing: the key is queried at every position rather than once per request ($+$36\% over a faithful request-level re-implementation), and its chains are merged into that tree rather than selected between.
On API-Bank the result is 4.4$\times$ autoregressive decoding speed (Llama-3.1-8B), the fastest drafter in our harness, trained or not, and the key adds most on hosts that mine the store less well.

\paragraph{Contributions.}
\begin{itemize}[leftmargin=1.05em, itemsep=0pt, topsep=1pt, parsep=0pt]
\item A method-independent diagnosis over ten workloads and two model families, separating what an
exact-suffix key cannot reach from what is simply absent.
\item A semantic draft source: a per-position hidden key, captured free during verification, over
the store a suffix drafter already keeps.
\item A merge that lets it ride inside an incumbent tree at that tree's own node budget; at matched
pool and budget it adds $+$24--29\% accepted length inside three published
drafters (Figure~\ref{fig:substrate}).
\end{itemize}

\section{Background and Related Work}
\label{sec:related}

\paragraph{Speculative decoding.} Write $x_1,x_2,\dots$ for the tokens a target model emits and
$t$ for the current step. Decoding one token normally costs one forward pass. At the batch
sizes this paper measures, that pass spends more time moving weights than doing arithmetic, so a
pass that scores many candidate positions at once costs little more than one that scores a single
position. Speculative decoding buys that headroom: the drafter proposes a continuation and the verifier commits the longest prefix
it agrees with. That makes the drafter the lever on $\tau$, since a proposal is either right or it
truncates the commit at its first wrong token.

\paragraph{Drafters, trained and training-free.} A \emph{trained} drafter learns to predict the target's next
tokens, from extra decoding heads on the target itself \citep{cai2024medusa} to EAGLE and its
successors, the strongest, which run a small autoregressive head on the target's own hidden
features, with EAGLE-3 fusing features from several depths rather than the top
layer alone \citep{li2024eagle,li2024eagle2,li2025eagle3}. A \emph{training-free} drafter instead
copies text it has already seen: prompt lookup decoding (PLD) from the current prompt
\citep{saxena2023pld}; SuffixDecoding from a suffix index over earlier responses, taking the
continuation of the longest repeated span \citep{oliaro2024suffixdecoding}; SAM-Decoding from a
suffix automaton over an offline corpus and over the current request at once
\citep{hu2025samdecoding}; REST from an offline corpus \citep{he2024rest}; Token Recycling from an adjacency table it fills for free with the top-$k$
candidates each verification pass has already computed and would otherwise discard
\citep{luo2024recycling}; and Lookahead from $n$-grams the model generates in parallel
\citep{fu2024lookahead}.

The two trade the same costs in opposite directions. A trained drafter proposes more accurately, but
needs a head trained for its target and spends a forward pass of its own per draft level. A
training-free one costs nothing to prepare and spends only a lookup, so it can be switched on for
any model on the day it ships; what it cannot do is invent, proposing only what its store already
contains, which makes it strongest exactly where serving traffic repeats itself.

A drafter need not propose a single
chain. Because the verification pass scores whatever it is given, it can score a \emph{tree} of
alternatives at once, using an attention mask that lets every node see only its own ancestors
\citep{miao2024specinfer,chen2024sequoia}; the pass then commits the longest path the target
agrees with. A tree costs what its nodes cost, so a system fixes a node budget per step and
decides what fills it. That opens a question a single-chain drafter never faces: if several
proposal sources are available, does one choose between them, or merge them? Merging is the
stronger answer, because two sources that agree on a prefix share its nodes and pay only where
they diverge, and it is established practice: RASD, Graft, RACER and READER all combine a
retrieval branch with another proposer inside one verified tree
\citep{quan2025rasd,graft2026,racer2025,reader2025}. We build on GOOSE, the strongest multi-source
tree decoder we measure \citep{goose2026}, which merges two training-free sources under one budget.

Every training-free drafter above is the same
mechanism: a \emph{pool} $\mathcal{P}$ of token sequences it may copy, and a \emph{key} that
decides which positions of that pool the current context retrieves. The key they share is
\emph{lexical}: match the longest suffix of the context that occurs verbatim in $\mathcal{P}$, back
off to shorter suffixes when it misses, and propose what followed the matches. Write $\mathcal{C}_{\mathrm{lex}}(t)$
for the set of token sequences that key proposes at step $t$. The stores differ---the current
request, a cache of past query-output pairs \citep{yang2023llma}, a suffix store over earlier
responses, an offline corpus---but the key does not, and it has one property that matters here: a
position is reachable only if its text recurs.

\paragraph{Semantic keys.} A \emph{semantic} key replaces the text with the model's own state:
the query is $h_t$, the state the target computed at step $t$; it retrieves the pool positions whose
stored states lie nearest $h_t$ in cosine similarity, above a floor $\theta$; and
$\mathcal{C}_{\mathrm{sem}}(t)$ is the corresponding set of proposals, the tokens that followed
those positions. Such keys begin with kNN-LM
\citep{khandelwal2020knnlm} and its descendants
\citep{he2021efficient,wang2023knnlm,drozdov2022cant,alon2022retomaton,li2024nearest}.

Their record in speculative decoding is weaker than the lexical key's. A dense match is an approximation. Two states can be near without their futures agreeing, so a system that lets it \emph{decide} anything pays for being wrong: SENSE relaxes the
acceptance rule and gives up losslessness, as does a wider training-free family
\citep{sense2026,loosesd2026,relaxedsd2026}, and SemanticSpec probes the same state to choose what
to accept \citep{semanticspec2026}. Systems that keep it narrow the key instead. DReSD is
the closest, with per-position hidden keys and strict losslessness, but it retrieves from a static offline
corpus and its key \emph{replaces} the lexical one rather than joining it \citep{dresd2025}.
ToolSpec, and an agent-serving co-design over a cross-session pool, coarsen the query to one vector
per request, selecting whole past traces and drafting inside them by exact match
\citep{toolspec2026,agentinfer2025}. AdaPLD keys its fallback on a static token embedding
\citep{adapld2026}, RACER and LogitSpec answer the same breakage with a logits source
\citep{racer2025,logitspec2026}, and ReSpec tunes when the lexical key fires rather than adding a
second one \citep{respec2025}. What none establishes is what a per-position semantic key is worth
\emph{beside} a strong lexical one over the same online pool, and that is what \S\ref{sec:gap}
measures.

\section{The Identifiability Gap}
\label{sec:gap}
%
%
\begin{figure}[t]\centering
\includegraphics[trim=46bp 13.4bp 24.9bp 22.3bp, clip, width=\columnwidth]{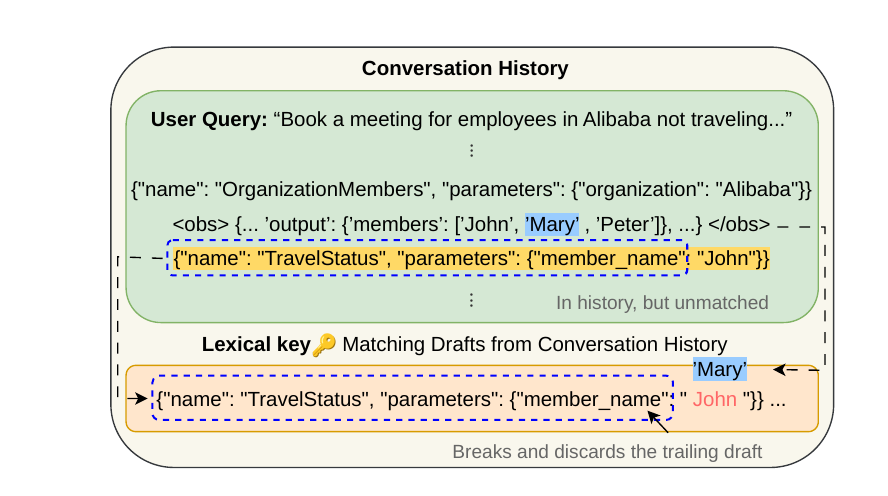}
\caption{\textbf{The answer is in the conversation, and exact matching still cannot reach it.} One
real API-Bank request, walked through in \S\ref{sec:gap}; the whole prompt and output are in
Figure~\ref{fig:case-toolspec-apibank}. \colorbox{fig1Blue}{\textbf{Blue}} marks the value the model is
about to generate, already present in the conversation;
\textcolor{cbRed!85!black}{red} marks what copying returns instead; the dashed box is the span
the lexical key matched and copied forward.}
\label{fig:gapproblem}
\end{figure}

Figure~\ref{fig:gapproblem} shows the failure in miniature, on one real API-Bank request.
The user has asked which colleagues are not travelling; the conversation has listed
\texttt{['John', 'Mary', 'Peter']} and checked John, so the model must now emit
\texttt{\{"member\_name": "Mary"\}}. The frame in front of that value, the opening of a
\texttt{TravelStatus} call, recurs verbatim from the John call a few lines up, so an exact-suffix
key matches it and copies that call forward, value included. It returns \texttt{John}. The cost is
not the one wrong token: verification commits only the matching prefix, so the closing
\texttt{"\}\}} the copy already held, and that was right, goes with it and must be bought again.

When a copying drafter fails there are exactly two possibilities. Either the
continuation is nowhere in the traffic the drafter may copy from, a \emph{coverage} failure no
copy method could repair, or it is there and the key cannot address it. Agent traffic supplies
both: this benchmark serves one conversation turn by turn, so a value minted by one request is a value
a later request asks for again; by then it is in the pool, with the same frame in front of it. Which dominates has not
been measured, and the two diagnoses recommend opposite fixes: a larger store, or a different key.

To settle it we measure the traffic itself, with no drafter in the loop. We replay a benchmark
request by request and stop at every token the model generated, asking of each what a server would
have had in its store at that moment: everything the requests before it contain. Is the continuation present at all, that is, do the next two tokens
occur verbatim in an earlier request? On API-Bank, at 92\% of positions they do. Is it addressable by
exact match? We put that to an \emph{exact-match oracle}, the strongest \emph{exact-match} lexical
drafter we can build over the same pool: it takes the longest suffix of the context that recurs, up
to sixteen tokens, backs off until it has eight candidates rather than one, and counts as right if
any of the eight is correct. It misses 13\% of positions all the same, and about half of those misses
were in the pool all along: the \emph{identifiability gap}, 6.8\% of generated positions, each with
a 6.4-token verbatim run behind it that a neighbour can match. That 6.8\% is not an artefact of how loosely
``present'' is defined: a four-token verbatim future instead of two drops availability from 92 to
80\% but the gap only from 6.8 to 5.2\%, and the key below reaches 88\% of it.

Why can an exact key not reach them? Because one token of the context is new, and every suffix
the key could match ends at that token: shortening the suffix does not step around it, so backing
off finds nothing either, down to the two-token key it stops at. The measurement shows that as a
cliff rather than a slope. Where the key succeeds it is matching sixteen tokens of context; at gap
positions the longest surviving match is one (Table~\ref{tab:mech}).

A semantic key does reach them, because what it matches is the situation and not the token. Its
query is the hidden state the verifier has already computed, so a past position in the same place of
the same kind of call is a near neighbour of this one even when it named a different colleague.
Queried at every position, on the oracle's own eight-candidate budget, it finds the true next token
at 81\% of the gap positions. The state is what does that: keyed instead on a static token
embedding, the model's input embedding for the token, which carries no context, the same
procedure reaches 23\%. The exact keys, for their part, largely reach the same positions as one
another; what ours reaches is different (Table~\ref{tab:venn}).
Tables~\ref{tab:mech} and~\ref{tab:census-grid} replicate the diagnosis on both families and ten
benchmarks, where what sits behind a recovered position differs sharply by traffic
(\S\ref{sec:setup}).

\section{\ours{}: a Semantic Source, and How It Merges into the Tree}
\label{sec:method}

\begin{figure*}[t]\centering
\includegraphics[width=\textwidth,trim=19.8bp 8.9bp 20.1bp 8.6bp,clip]{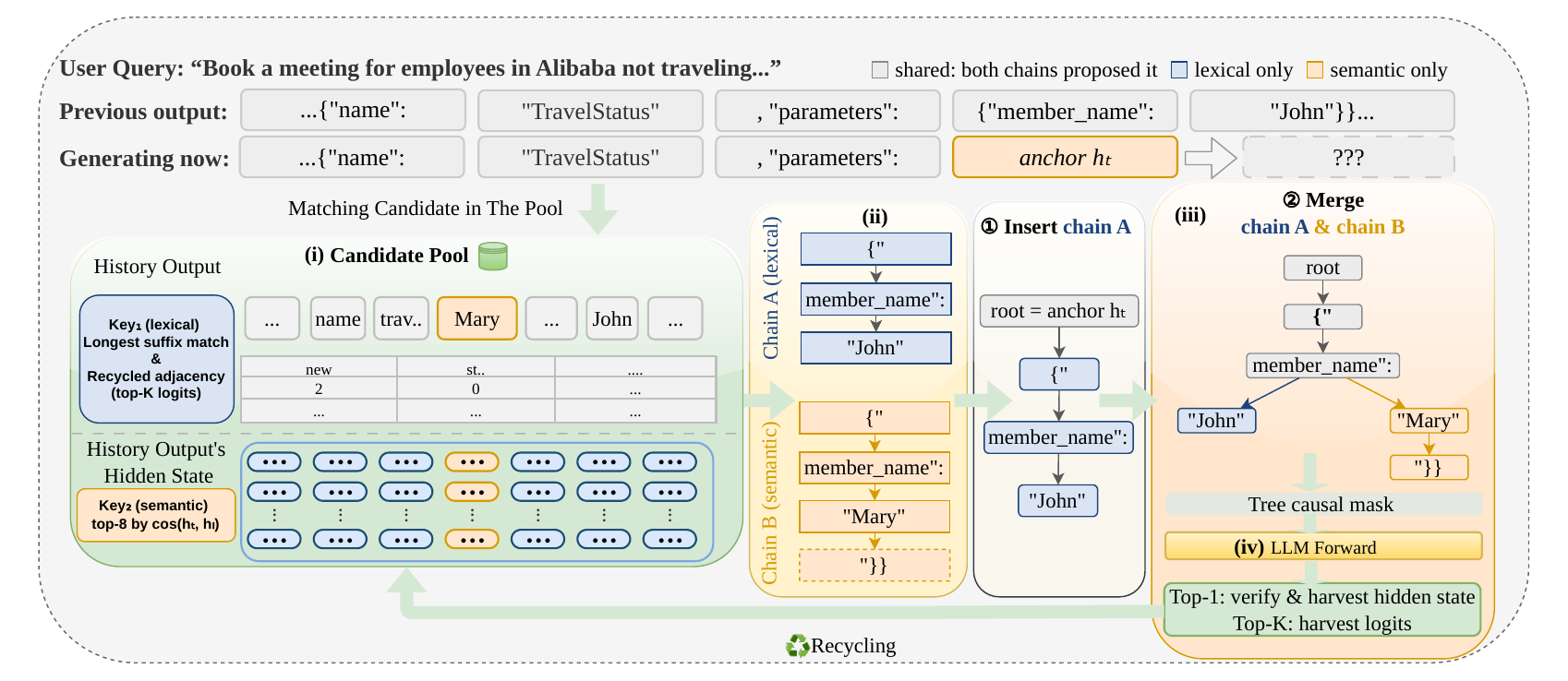}
\caption{\textbf{\ours{}: one store, two keys, one tree.} One drafting cycle on a request like
Figure~\ref{fig:gapproblem}'s, at the point where the value it needs has been emitted before and is
in the pool: the identifiability gap of \S\ref{sec:gap}.
\textbf{(i)~Store}: one record per past position, the token and the verifier's hidden state
$(x_i,h_i)$, with a longest-suffix index on the tokens and a cosine index on the states; beside them
sits the incumbent's own by-product of the same forward, the token-adjacency table of Token
Recycling \citep{luo2024recycling}, which we leave as it is.
\textbf{(ii)~Retrieve}: the lexical key runs first and the semantic key only where that match is
non-empty. \textbf{(iii)~Merge}: the semantic chain re-walks the lexical path from the root, so a
node whose token is already there is walked rather than bought. \textbf{(iv)~Verify}: one forward
covers the whole tree and is the same forward that computes the states of (i). Algorithm~\ref{alg:step} is the step as deployed.}
\label{fig:pipeline}
\end{figure*}

\ours{} adds one draft \emph{source}: the proposals $\mathcal{C}_{\mathrm{sem}}$ of the semantic
key of \S\ref{sec:related}, filling part of a multi-source draft tree beside the two lexical
sources that fill the rest---a suffix automaton over the store, and a table of token adjacencies.
Figure~\ref{fig:pipeline} draws one drafting cycle in four stages---store, retrieve, merge,
verify---and Algorithm~\ref{alg:step} is the same cycle line by line.

\begin{algorithm}[t]
\small
\caption{One \ours{} drafting step, as deployed; \S\ref{sec:method} walks it and
Table~\ref{tab:repro} configures it.}
\label{alg:step}
\begin{algorithmic}[1]
\Require $\mathcal{P}$, $h_t$, $\mathcal{C}_{\mathrm{lex}}$, $D$ as in
  Eq.~(\ref{eq:store})--(\ref{eq:copy}); $k{=}8$ neighbours, $\theta{=}0.8$, layer
  $\mathrm{round}(0.85L)$ of the model's $L$ layers, $B_{\mathrm{ret}}{=}16$ of $60$ nodes
\If{$\mathcal{C}_{\mathrm{lex}}=\emptyset$ \textbf{or} the anchor has no adjacency successors}
  \Return{adjacency-only draft, a chain verify of $\mathcal{C}_{\mathrm{lex}}$ alone, or one AR
    step, whichever is available} \label{alg:gate}
\EndIf
\State $T\gets\textsc{BuildLexical}(\mathcal{C}_{\mathrm{lex}},\,60{-}B_{\mathrm{ret}})$ \label{alg:lex} \Comment{(ii) retrieve}
\State $s_i\gets\cos(h_t,h_i)\;\forall i\in\mathcal{P}$;\;\, $E_t\gets\textrm{top-}k(s)$;\;\,
  $\mathcal{C}\gets\emptyset$ \Comment{candidate chains}
\If{$\max_i s_i\ge\theta$ \textbf{and} \textsc{SelfCal} not muting}
  \ForAll{$i\in E_t$ by decreasing $s_i$}
    \State $d\gets D(s_i)$;\; \textbf{if} $d{=}0$ \textbf{continue}
    \State $\gamma\gets x_{i+2:\,i+1+d}$ \label{alg:offset}
    \State \textbf{if} $|\gamma|{\ge}2$ \textbf{and} $\gamma\notin\mathcal{C}$:\;
      $\mathcal{C}\gets\mathcal{C}\cup\{\gamma\}$
  \EndFor
\EndIf
\State $\mathcal{C}\mathrel{\cup}=$ top-$3$ automaton branches by frequency, $\le32$ tokens \Comment{lexical}
\State $\mathcal{C}\mathrel{\cup}=$ $\le4$ rejected-branch rollouts, anchor $x$ equal and
  $\cos{\ge}0.8$ \label{alg:recycle} \Comment{recycled pool}
\State $T\gets\textsc{MergeFromRoot}(T,\gamma)$ for $\gamma\in\mathcal{C}$, until
  $B_{\mathrm{ret}}$ spent \Comment{(iii) merge}
\State $y\gets\textsc{VerifyMaskedGreedy}(T)$; commit the longest accepted path \emph{and} the bonus token \Comment{(iv) verify}
\State $h_t\gets$ final hidden state; append the trace to $\mathcal{P}$ at request end \Comment{(i) store}
\end{algorithmic}
\end{algorithm}
\paragraph{(i) Store.} One record per position committed inside a
verifying forward,
\begin{equation}\label{eq:store}
\mathcal{P}=\{(x_i,h_i)\},\quad h_i\in\mathbb{R}^{d_{\mathrm{model}}}\ \text{at layer}\ \ell,
\end{equation}
each $h_i$ a real vector of the model's width $d_{\mathrm{model}}$, read at layer $\ell=\mathrm{round}(0.85L)$ of its $L$ layers and kept normalized in half precision ($\approx$8\,KB per token at 8B). Capture costs a copy, never a forward pass, so the store fills by serving. A request's trace enters the store only once that request has
finished, and only the tokens the model generated; prompts are never stored. Retrieval skips any
stored trace whose own prompt is the one being served now, so a request cannot draft from a copy of
itself. Two indexes sit over that one store, which is all that ``one store, two keys'' means: a
longest-suffix index over its tokens, and a cosine index over the positions carrying a state.

\paragraph{(i\,b) Recycling.} Token Recycling makes a store out of what verification throws away:
the top-$k$ tokens the forward has already computed at every node of the draft tree, kept in an
adjacency table keyed by the token alone \citep{luo2024recycling}. The same forward also computes a
hidden state at every node, and that is discarded. We keep it, for the nodes verification
\emph{rejects} as well as those it accepts. An accepted node's state is a record of (i); a rejected
one becomes a record keyed by the state before its token and the token itself, whose value is the
continuation the model endorses from there---read out of the same forward by following the child
whose drafted token is that node's own argmax, up to six deep. These live in a ring buffer of their own, beside the trace store
(line~\ref{alg:recycle}): that store holds the long continuations actually generated, this one
the short ones endorsed but never committed.

\paragraph{(ii) Retrieve.} Every source keys off the \emph{anchor}: the last committed token,
the root of the tree being built. For the lexical half of that tree we use the standard machinery
unchanged (line~\ref{alg:lex}): a suffix automaton over the store \emph{and} the current request's
own committed tokens, which contributes its longest match's continuation and its next most frequent
branches, and Token Recycling's token-adjacency table, which fills what is left. The semantic key fires only where that lexical
match is non-empty (line~\ref{alg:gate}): it extends a chain that exists rather than opening one. Where it
does fire, it is queried at \emph{every} position rather than once per request, over the same pool
under the other index. Writing $s_i=\cos(h_t,h_i)$ for the similarity to a stored state, the entries
it retrieves are
\begin{equation}\label{eq:retr}
E_t=\operatorname*{arg\,top\text{-}\textit{k}}_{i\in\mathcal{P}}\ s_i
\quad\text{if}\ \ \max_{i\in\mathcal{P}} s_i\ge\theta ,
\end{equation}
so $E_t$ holds the $k{=}8$ pool positions of highest similarity and $\theta$ is a firing condition on
the best of them, not a filter inside the top-$k$. One similarity per position lets each retrieved
$i$ set its own copy depth, in tokens: the chain $\gamma_i$ it contributes is read straight out of
the trace $x$ that holds it,
\begin{align}
\gamma_i &= x_{\,i+2\,:\,i+1+D(s_i)}, \label{eq:copy}\\[-2pt]
D(s) &= 48\ \text{if}\ s\!\ge\!0.90;\ 8\ \text{if}\ s\!\ge\!0.82;\ 0\ \text{otherwise} \nonumber
\end{align}
which a request-level key cannot do, having one similarity per trace (\S\ref{sec:granularity}).
The query state is the one the last \emph{verifying} pass left, so it lags the drafted position by
a token or more, uncorrected; that lag is why Eq.~(\ref{eq:copy}) copies from two positions after
the neighbour, not one.
Self-calibration (\textsc{SelfCal} in Algorithm~\ref{alg:step}) stops querying the key
where its drafts are rarely accepted.

\paragraph{(iii) Merge.} Each source returns a linear chain, and one tree $T$ has to hold them all
under one node budget. Hanging each under the root double-pays every token two sources
agree on, and agreement is common. Instead each retrieved chain \emph{re-walks} the tree
from the root: wherever the next token is already carried by a child, that child is walked rather
than bought, so a chain $\gamma$ costs
\begin{equation}\label{eq:cost}
\mathrm{cost}(\gamma,T)=|\gamma|-\pi(\gamma,T),
\end{equation}
where $\pi(\gamma,T)$ is the length of the longest prefix of $\gamma$ already present as a root
path in $T$. A token two sources propose is paid for once, not twice. Retrieved chains therefore join the tree part-way
down rather than at the root: the lexical sources have already bought the root and the tokens under
it, so a retrieved chain buys nodes only from where it stops agreeing with them. That is where the
saving comes from, and it is not a stylistic choice: made to hang from the root instead, at the
same budget, the same chains cost more than the source was measured to add
(Table~\ref{tab:arms}). A node added this way attends to itself
and its ancestors and to nothing else, so the merge only \emph{adds} branches and leaves the
acceptance rule untouched. The retrieved chains spend a fixed reservation, $B_{\mathrm{ret}}{=}16$
of the tree's 60 nodes, taken out of the lexical allocation rather than added to it, so
verification cost does not move.

\paragraph{(iv) Verify.} One forward covers the whole tree, masked so that every node sees only
its own ancestors and verified greedily.
The longest accepted path is committed \emph{together with} the \emph{bonus token} that forward has
already computed at its last node---when nothing is accepted, that token alone. That same forward supplies (i)'s hidden states and the recycled pool's rejected
nodes: the store is filled by the work of serving, not by a pass of its own. Verification is unchanged and greedy, and no part of its decision moves forward into drafting.

\section{Experimental Setup}
\label{sec:setup}

\paragraph{Models.}
We use Llama-3.1-8B-Instruct and Qwen3-8B in bfloat16 under greedy decoding. The scale ladder stays within Qwen3, at 8B, 14B and 32B (Table~\ref{tab:main-echo}), so that a change down the column is a change in model size and nothing else.

\paragraph{Datasets.}
Each benchmark is served in order, the pool accumulating across its requests. Eight: API-Bank
\citep{li2023apibank}, ToolAlpaca \citep{tang2023toolalpaca}, tau-bench retail \citep{yao2024tau},
HumanEval \citep{chen2021codex}, ClassEval \citep{du2023classeval}, GSM8K \citep{cobbe2021gsm8k},
MT-Bench \citep{zheng2023mtbench} and WildChat \citep{zhao2024wildchat}. The first three are the
repetitive traffic we target; the rest test for regressions. The diagnosis of \S\ref{sec:gap} adds tau-bench
airline and Spec-Bench math \citep{xia2024specbench}, ten benchmarks in all. Sizes and generation
caps are in
Table~\ref{tab:matrix}.

\begin{table*}[t]
\centering\footnotesize
\setlength{\tabcolsep}{2.15pt}
\begin{tabular}{@{}l rr rr rr rr @{\hskip 9pt} rr rr rr rr@{}}
\toprule
& \multicolumn{8}{c}{\textit{Llama-3.1-8B}} & \multicolumn{8}{c@{}}{\textit{Qwen3-8B}} \\
\cmidrule(lr){2-9}\cmidrule(l){10-17}
& \multicolumn{2}{c}{API-Bank} & \multicolumn{2}{c}{ToolAlpaca} & \multicolumn{2}{c}{tau-Retail} & \multicolumn{2}{c}{\textbf{Avg.}}
& \multicolumn{2}{c}{API-Bank} & \multicolumn{2}{c}{ToolAlpaca} & \multicolumn{2}{c}{tau-Retail} & \multicolumn{2}{c@{}}{\textbf{Avg.}} \\
Method & Spd & $\tau$ & Spd & $\tau$ & Spd & $\tau$ & Spd & $\tau$ & Spd & $\tau$ & Spd & $\tau$ & Spd & $\tau$ & Spd & $\tau$ \\
\midrule
AR & 1.00$\times$ & 1.00 & 1.00$\times$ & 1.00 & 1.00$\times$ & 1.00 & 1.00$\times$ & 1.00 & 1.00$\times$ & 1.00 & 1.00$\times$ & 1.00 & 1.00$\times$ & 1.00 & 1.00$\times$ & 1.00 \\
GOOSE & 2.37$\times$ & 3.18 & 2.17$\times$ & 2.73 & 1.47$\times$ & 2.08 & 2.00$\times$ & 2.66 & 1.98$\times$ & 2.67 & 1.64$\times$ & 2.05 & 1.22$\times$ & 1.86 & 1.61$\times$ & 2.19 \\
Token Recycling & 1.94$\times$ & 2.70 & 1.92$\times$ & 2.60 & 1.50$\times$ & 2.30 & 1.79$\times$ & 2.53 & 1.66$\times$ & 2.36 & 1.71$\times$ & 2.37 & 1.29$\times$ & 2.34 & 1.55$\times$ & 2.36 \\
SuffixDecoding & 3.88$\times$ & 4.87 & 2.15$\times$ & 2.35 & 1.94$\times$ & 2.67 & 2.66$\times$ & 3.30 & 3.49$\times$ & 4.51 & 2.26$\times$ & 2.57 & 1.68$\times$ & 2.97 & 2.48$\times$ & 3.35 \\
\ours{} (Ours) & \textbf{4.39$\times$} & \textbf{6.76} & 2.61$\times$ & 3.44 & \textbf{2.22$\times$} & \textbf{4.06} & \textbf{3.07$\times$} & \textbf{4.75} & \textbf{3.95$\times$} & \textbf{6.33} & \textbf{2.60$\times$} & 3.57 & \textbf{1.83$\times$} & \textbf{4.45} & \textbf{2.79$\times$} & \textbf{4.78} \\
\rowcolor{gray!15}
EAGLE-3 \emph{(trained)} & 2.03$\times$ & 3.81 & \textbf{2.93$\times$} & \textbf{5.09} & 1.10$\times$ & 3.09 & 2.02$\times$ & 4.00 & 1.68$\times$ & 3.23 & 1.98$\times$ & \textbf{3.61} & 1.01$\times$ & 3.15 & 1.56$\times$ & 3.33 \\
\midrule
& \multicolumn{8}{c}{\textit{Qwen3-14B}} & \multicolumn{8}{c@{}}{\textit{Qwen3-32B}} \\
\cmidrule(lr){2-9}\cmidrule(l){10-17}
AR & 1.00$\times$ & 1.00 & 1.00$\times$ & 1.00 & 1.00$\times$ & 1.00 & 1.00$\times$ & 1.00 & 1.00$\times$ & 1.00 & 1.00$\times$ & 1.00 & 1.00$\times$ & 1.00 & 1.00$\times$ & 1.00 \\
GOOSE & 2.03$\times$ & 2.65 & 1.74$\times$ & 2.15 & 1.24$\times$ & 1.81 & 1.67$\times$ & 2.20 & 2.32$\times$ & 2.85 & 1.89$\times$ & 2.12 & 1.30$\times$ & 1.77 & 1.84$\times$ & 2.25 \\
Token Recycling & 1.71$\times$ & 2.32 & 1.67$\times$ & 2.22 & 1.35$\times$ & 2.37 & 1.58$\times$ & 2.30 & 2.09$\times$ & 2.60 & 1.91$\times$ & 2.31 & 1.47$\times$ & 2.36 & 1.82$\times$ & 2.42 \\
SuffixDecoding & 3.45$\times$ & 4.47 & 1.82$\times$ & 2.03 & 1.68$\times$ & 2.97 & 2.32$\times$ & 3.16 & 4.17$\times$ & 4.68 & 2.03$\times$ & 2.00 & 2.05$\times$ & 3.16 & 2.75$\times$ & 3.28 \\
\ours{} (Ours) & \textbf{4.08$\times$} & \textbf{6.33} & \textbf{2.31$\times$} & 3.05 & \textbf{1.80$\times$} & \textbf{4.12} & \textbf{2.73$\times$} & \textbf{4.50} & \textbf{4.51$\times$} & \textbf{6.15} & \textbf{2.41$\times$} & 2.96 & \textbf{2.09$\times$} & \textbf{4.24} & \textbf{3.00$\times$} & \textbf{4.45} \\
\rowcolor{gray!15}
EAGLE-3 \emph{(trained)} & 2.15$\times$ & 4.09 & 1.82$\times$ & \textbf{3.12} & 1.05$\times$ & 3.00 & 1.67$\times$ & 3.40 & 2.21$\times$ & 3.64 & 2.23$\times$ & \textbf{3.36} & 1.07$\times$ & 2.93 & 1.84$\times$ & 3.31 \\
\bottomrule
\end{tabular}
\caption{\textbf{The three repetitive workloads.} Wall-clock speedup over autoregressive decoding (Spd)
and accepted length ($\tau$), entire benchmark, greedy; autoregressive decoding is the reference for
both and is the 1.00 row. The 32B block carries its own autoregressive reference, re-measured
beside it (Table~\ref{tab:hardware}). \textbf{Bold} marks the best cell in each column; the trained row is shaded, and it takes
ToolAlpaca. Avg.\ is
the mean of the three workloads. EAGLE-3's Llama API-Bank cell is the full 597-request benchmark;
its Qwen API-Bank cells are the benchmark's first 100 requests and its tau-Retail cells the 46--54
of 69 inside its 4096-token KV cap, with its speedup divided by the autoregressive baseline
restricted to the same requests. Speedups read within a family block, never down it.}
\label{tab:main-echo}
\end{table*}
A second run is instrumented to record which source delivered each accepted token. It keeps the
three repetitive workloads of Table~\ref{tab:main-echo} and adds eight further benchmarks, spanning
highly repetitive traffic to barely repetitive at all
\citep{li2021mtop,valmeekam2023planbench,yu2018spider,suzgun2023bbh,jimenez2024swebench,wei2024magicoder,barres2025tau};
$\tau^2$-bench contributes two domains, for eleven in all at 50--198 requests each. Three of the
eleven are restricted to a single domain, homogeneity being the property under test.

\paragraph{Baselines.} Eight, each measured where it is defined: AR; GOOSE \citep{goose2026}, the
fusion tree \ours{} adopts; EAGLE-3 \citep{li2025eagle3} as the trained reference; and SuffixDecoding and Token Recycling
\citep{oliaro2024suffixdecoding,luo2024recycling}, re-run in our harness at the node budget and
verifier of Table~\ref{tab:matrix}. ToolSpec \citep{toolspec2026} is not re-run here; it appears in
Figure~\ref{fig:substrate} as a transplant host, measured inside its own released harness. PLD \citep{saxena2023pld} and Lookahead
\citep{fu2024lookahead} trail GOOSE on every workload we tried; REST
\citep{he2024rest} and CREST \citep{crest2024} are not run, their online analogue being the
SuffixDecoding arm.

\paragraph{Metrics.}
Accepted length $\tau$, committed tokens per verification pass, pooled per dataset; and wall-clock
speedup over autoregressive decoding in the same harness. A draft
token is accepted only where it equals the token the model would emit after that node's in-tree
prefix, so every arm is lossless with respect to greedy decoding, which we verify token for token
rather than assert (Table~\ref{tab:repro}). The main tables are measured at batch size one; a
separate probe measures what changes at batch sizes up to 32 (Figure~\ref{fig:batchprobe}). Table~\ref{tab:hardware}
gives the hardware and software configuration; accepted length and wall-clock are read
within one machine and never across it. Pools are strictly
online, with leakage excluded, and an audit script recomputes 765 of
this paper's numbers from raw data.

\section{Results and Analysis}
\label{sec:results}
\label{sec:analysis}

\label{sec:results-main}
\paragraph{Main results.}
\ours{} takes the repetitive workloads of Table~\ref{tab:main-echo}. On API-Bank with
Llama-3.1-8B it commits $\tau$ 6.76 tokens per verification pass at 4.39$\times$ autoregressive
decoding speed, against 4.87 at 3.88$\times$ for SuffixDecoding, the strongest training-free
baseline, 3.18 at 2.37$\times$ for GOOSE, the tree it drafts inside, and 3.81 at 2.03$\times$ for
the trained EAGLE-3. Averaged over the three workloads it leads on both metrics in all four model
blocks: $\tau$ 4.75 at 3.07$\times$ on Llama-3.1-8B, 4.78 at 2.79$\times$ on Qwen3-8B, 4.50 at 14B
and 4.45 at 32B. The one exception is ToolAlpaca, where the trained EAGLE-3 takes accepted length.

\paragraph{Batching.}
Speculation's advantage over autoregressive decoding at the same batch size falls from
1.3--1.6$\times$ at batch one to 0.8--1.1$\times$ at batch 32 on API-Bank
(Figure~\ref{fig:batchprobe}), and the largest batch that still pays shrinks as the tree grows. The semantic key does not
survive that regime either. Queried per position, as in deployment, it moves accepted
length by $-$2.4 to $+$2.7\% across batch sizes 1 to 32---inside the harness's own repeat
noise, and per node bought it is negative in 19 of the 20 cells.

\paragraph{End-to-end speedup.}
Serving pays in wall-clock, not $\tau$: on API-Bank \ours{} is the fastest drafter in this harness,
trained or not, at 4.39$\times$ (Llama) end-to-end, SuffixDecoding next at 3.88$\times$
(Table~\ref{tab:main-echo}). Off the repetitive workloads the semantic source is neutral rather
than costly, and \ours{} still leads the training-free field on both families
(Table~\ref{tab:main-8bench}). Off that traffic a trained drafter is the better choice: EAGLE-3
leads us on every one of those five workloads. Accepted length holds up the Qwen3 ladder: on
API-Bank $\tau$ 6.33 at 8B and at 14B, 6.15 at 32B (Table~\ref{tab:main-echo}).

\begin{figure}[t]\centering
\def\abLadj{-12.7} \def\abLadjE{0.15}   \def\abQadj{-8.5}  \def\abQadjE{0.70}
\def\abLaut{-10.6} \def\abLautE{0.82}   \def\abQaut{-14.4} \def\abQautE{0.65}
\def\abLkey{-3.5}  \def\abLkeyE{0.25}   \def\abQkey{-4.5}  \def\abQkeyE{0.85}
\def\tbLadj{-28.4} \def\tbLadjE{1.03}   \def\tbQadj{-22.9} \def\tbQadjE{0.21}
\def\tbLaut{-20.5} \def\tbLautE{0.00}   \def\tbQaut{-21.7} \def\tbQautE{0.44}
\def\tbLkey{-6.4}  \def\tbLkeyE{1.35}   \def\tbQkey{-4.3}  \def\tbQkeyE{0.12}
\begin{tikzpicture}
\pgfplotsset{
  keypanel/.style={
    scale only axis, width=2.20cm, height=2.10cm, font=\scriptsize,
    axis line style={black!55}, tick align=outside, tick style={black!55},
    xmajorgrids, grid style={black!12, line width=0.35pt}, axis on top,
    xbar, bar width=4.0pt, clip=false,
    xmin=-34, xmax=0.6, xtick={-30,-20,-10,0},
    xticklabel style={font=\tiny},
    ymin=0.45, ymax=3.55, ytick={1,2,3},
    title style={font=\scriptsize, yshift=-4pt},
    error bars/x dir=both,
    error bars/x explicit,
    error bars/error bar style={black!55, line width=0.35pt},
    error bars/error mark options={black!55, mark size=1.1pt, line width=0.35pt},
  },
  llamabar/.style={fill=gkNavy, draw=gkNavy, bar shift=2.6pt},
  qwenbar/.style={fill=gkOrange, draw=gkOrange!80!black, bar shift=-2.6pt},
}
\newcommand{\vlab}[4]{\node[anchor=east, font=\tiny, text=#3, yshift=#2, inner sep=0pt]
  at (axis cs:12.6,#1) {$#4$};}
\begin{axis}[name=ab, keypanel, title={API-Bank},
    yticklabels={$-$\,semantic\\key, $-$\,suffix\\automaton, $-$\,token\\adjacency},
    yticklabel style={font=\scriptsize, align=right}]
  \addplot[llamabar] coordinates
    {(\abLkey,1) +- (\abLkeyE,0)  (\abLaut,2) +- (\abLautE,0)  (\abLadj,3) +- (\abLadjE,0)};
  \addplot[qwenbar] coordinates
    {(\abQkey,1) +- (\abQkeyE,0)  (\abQaut,2) +- (\abQautE,0)  (\abQadj,3) +- (\abQadjE,0)};
  \vlab{3}{2.6pt}{gkNavy}{\abLadj}   \vlab{3}{-2.6pt}{gkOrange!75!black}{\abQadj}
  \vlab{2}{2.6pt}{gkNavy}{\abLaut}   \vlab{2}{-2.6pt}{gkOrange!75!black}{\abQaut}
  \vlab{1}{2.6pt}{gkNavy}{\abLkey}   \vlab{1}{-2.6pt}{gkOrange!75!black}{\abQkey}
\end{axis}
\coordinate (rp) at ([xshift=0.92cm]ab.south east);
\begin{axis}[name=tb, keypanel, at={(rp)}, anchor=south west,
    title={tau-bench retail}, yticklabels={,,}]
  \addplot[llamabar] coordinates
    {(\tbLkey,1) +- (\tbLkeyE,0)  (\tbLaut,2) +- (\tbLautE,0)  (\tbLadj,3) +- (\tbLadjE,0)};
  \addplot[qwenbar] coordinates
    {(\tbQkey,1) +- (\tbQkeyE,0)  (\tbQaut,2) +- (\tbQautE,0)  (\tbQadj,3) +- (\tbQadjE,0)};
  \vlab{3}{2.6pt}{gkNavy}{\tbLadj}   \vlab{3}{-2.6pt}{gkOrange!75!black}{\tbQadj}
  \vlab{2}{2.6pt}{gkNavy}{\tbLaut}   \vlab{2}{-2.6pt}{gkOrange!75!black}{\tbQaut}
  \vlab{1}{2.6pt}{gkNavy}{\tbLkey}   \vlab{1}{-2.6pt}{gkOrange!75!black}{\tbQkey}
\end{axis}
\node[anchor=north, font=\scriptsize] at ([yshift=-11pt, xshift=0.46cm]ab.south east)
  {change in accepted length $\tau$, \%};
\node[anchor=north, font=\tiny] at ([yshift=-22pt, xshift=0.46cm]ab.south east)
  {\textcolor{gkNavy}{\rule{4.5pt}{4.5pt}}~Llama-3.1-8B\quad
   \textcolor{gkOrange}{\rule{4.5pt}{4.5pt}}~Qwen3-8B};
\end{tikzpicture}
\caption{\textbf{Accepted length lost when each draft source is removed.} One bar per model family,
on the two repetitive workloads. Every bar is negative: no source is free to remove. The semantic key is the
smallest of the three, both lexical sources outweigh the key everywhere (1.9--5.3$\times$). Each
API-Bank cell is that benchmark's opening 200 of 597 requests, each tau-bench retail cell all 69.}
\label{fig:keygrid}
\end{figure}
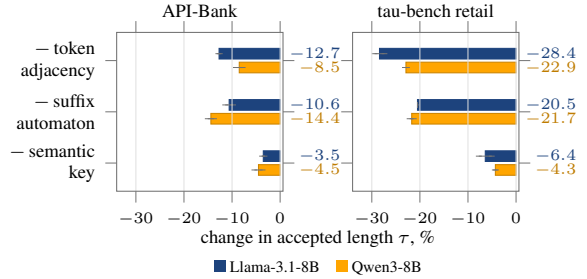

\subsection{Ablation}
\label{sec:granularity}
The system carries three draft sources; this section asks what each is for. We remove them one at a
time from the deployed system, on both model families and both repetitive workloads, and read what
accepted length is lost (Figure~\ref{fig:keygrid}).

Nothing is free to remove. A lexical source costs up to 28.4\% of accepted length and the semantic
key up to 6.4\%, and the key's loss is real in every cell, never zero and never favourable. It is also the smallest of the three, which is what the design predicts: the key is aimed at the
7\% of positions \S\ref{sec:gap} identifies, and the lexical sources carry the rest.

The key is an \emph{extender} rather than a drafter. With both lexical sources removed it stops drafting
altogether, committing exactly one token per verification pass: it can lengthen a chain that exists
but cannot open one. Its value is always marginal, and the right way to measure it is a
controlled pair: over the entire 597-request API-Bank benchmark, the deployed system against the
same system with only the semantic entry budget zeroed, the key is worth $+$6.2\% of accepted
length ($\tau$ 6.36 to 6.75, which three further node allocations reproduce within 0.2 points). That is its worth on the strongest lexical host; on weaker
hosts it is larger (\S\ref{sec:generality}).

What a recovery returns is the place, not the value: the unrepeatable content tokens are 31\% of what
our retriever recovers but 71\% of what no copy method reaches (Llama; 44 and 74\% on Qwen3). What the key contributes is re-synchronisation.
It puts the draft back onto the structure that resumes once the value has passed, and the value
arrives free one step later as the verifier's bonus token: on API-Bank a recovered position is
followed by a 6.8-token verbatim run, of which a neighbour matches 6.4 (Llama), while on math, code
and chat what the key finds is spent within a few tokens.

\paragraph{Retrieval granularity.} Two things separate our key from a request-level one,
and they are worth different amounts. Over a lexical-only base of $\tau$ 3.07, at a fixed pool, node budget and verifier, moving
the key from once per request to every position,
with the copy depth assigned by neighbour rank rather than by similarity, raises accepted length from $\tau$ 3.58 to
3.89 ($+$8.6\%). Letting the match's own similarity choose that depth, which a request-level key
cannot do with one similarity per trace rather than one per position, raises it again to 4.86
($+$25.0\%). Both are ours, and together they are the $+$36\% over a request-level
re-implementation; but the second term is the larger, and it is a consequence of the first rather
than an independent knob.

\paragraph{Component ablation.} No single component is load-bearing. Removing any one of similarity-sized depth, the entry
floor, self-calibration or the rejected-branch pool moves API-Bank accepted length by at most 2.7\%
on either family, and knocked out \emph{as a group} they cost 3.3--8.0\%: depth reaches the draft
from more than one of them at once (Table~\ref{tab:interaction}).

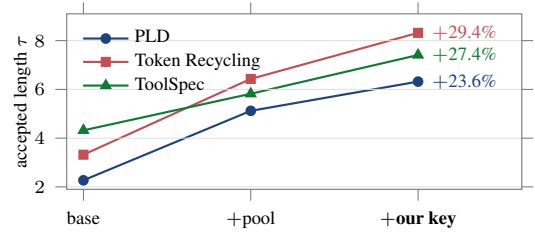
\begin{figure}[t]\centering
\begin{tikzpicture}
\begin{axis}[
    width=7.6cm, height=3.9cm, font=\scriptsize,
    axis line style={black!55}, tick align=outside, tick style={black!55},
    ymajorgrids, grid style={black!12, line width=0.35pt}, axis on top,
    ylabel={accepted length $\tau$}, ylabel style={font=\scriptsize, yshift=-6pt},
    xtick={0,1,2}, xticklabels={base, $+$pool, $+$\textbf{our key}},
    xticklabel style={font=\scriptsize}, xmin=-0.10, xmax=2.62,
    ymin=1.9, ymax=9.10, ytick={2,4,6,8},
    legend style={at={(0.015,0.97)}, anchor=north west, draw=none, fill=none,
                  font=\scriptsize, row sep=-1.5pt, legend cell align=left},
    clip=false,
  ]
  \addplot[gkNavy, thick, mark=*, mark size=1.6pt]
    coordinates {(0,2.27) (1,5.12) (2,6.32)};
  \addlegendentry{PLD}
  \node[anchor=west, font=\scriptsize, inner sep=2pt, text=gkNavy] at (axis cs:2.05,6.32) {$+$23.6\%};
  \addplot[cbRed, thick, mark=square*, mark size=1.5pt]
    coordinates {(0,3.32) (1,6.43) (2,8.32)};
  \addlegendentry{Token Recycling}
  \node[anchor=west, font=\scriptsize, inner sep=2pt, text=cbRed] at (axis cs:2.05,8.32) {$+$29.4\%};
  \addplot[gkGreen, thick, mark=triangle*, mark size=1.9pt]
    coordinates {(0,4.32) (1,5.82) (2,7.41)};
  \addlegendentry{ToolSpec}
  \node[anchor=west, font=\scriptsize, inner sep=2pt, text=gkGreen] at (axis cs:2.05,7.41) {$+$27.4\%};
\end{axis}
\end{tikzpicture}
\caption{\textbf{Three published drafters at three rungs, with our key transplanted into each.} The
rungs are: the drafter alone (\emph{base}), the same
drafter handed our cross-request store under its own exact-match key ($+$\emph{pool}), and that key
replaced by the semantic one at an unchanged node budget ($+$\emph{our key}); the percentage beside
each line is its gain across the last rung. }
\label{fig:substrate}
\end{figure}

\subsection{Generality}
\label{sec:generality}
\label{sec:whenpays}
The key is not an artefact of the system we built it in. Handed to three published training-free
drafters on the entire API-Bank benchmark, each already holding the same cross-request pool under
its own exact-match key, it lifts every one of them at an unchanged node budget: $+$23.6\% of
$\tau$ on prompt lookup, $+$29.4\% on Token Recycling and $+$27.4\% on ToolSpec
(Figure~\ref{fig:substrate}). The pool those hosts are handed is prior art
\citep{oliaro2024suffixdecoding,he2024rest}, worth $+$126, $+$94 and $+$35\% to them on its own;
the key is what re-keying that store is worth on top of it, on drafters we did not write.

Whether to enable the source is decided by one number, read off the instrumented run of
\S\ref{sec:setup}: the \emph{sole-source share}, the fraction of accepted tokens that no other
source delivered. Over the eleven benchmarks it runs from 0.9\% to 21.1\%, and it separates the
regimes: above ${\sim}9\%$ the semantic-only arm beats GOOSE on both accepted length and
throughput, below ${\sim}3\%$ it loses on both, and the deployed merged system is
neutral-to-positive throughout.

\section{Conclusion}
\label{pg:concl}

Training-free speculative decoding misses correct candidates that its own history already holds,
and the cause is addressing rather than coverage: on our densest tool-calling benchmark, about half of
what the strongest exact-match drafter misses is text already emitted, put out of reach by one
minted value.
\ours{} closes that gap with a second key over the store the drafter already keeps, the hidden
state the verifier computes anyway, merged into an existing draft tree at its own budget. It reaches 4.4$\times$ autoregressive decoding speed on API-Bank and lifts
accepted length by 24 to 29\% inside three drafters we did not write. The diagnosis is measured on the traffic, not on our
method.

\section*{Limitations}
\label{pg:lim}
\ours{} is task-dependent by design: it pays where traffic re-visits the same situations and is
neutral where it does not.

The semantic source is not free on either axis. It stores one hidden vector per generated token,
about 8\,KB at 8B scale, so a full 597-request API-Bank benchmark holds ${\approx}$0.27\,GB beside
the KV cache and our longest benchmark ${\approx}$2.6\,GB; keeping that store bounded in a
long-running deployment---by eviction, an index, or quantization---is future work.

A training-free drafter runs no model of its own at draft time, which is the cost a trained drafter
pays; what it pays instead is retrieval. Ours costs 4.8 to 5.8\% of a verification cycle, most of
it keeping the rejected-branch pool rather than the cosine lookup, which is why accepted length
converts to wall-clock at less than one for one. That fee is a limitation of retrieval-based
training-free drafting in general, not of this key alone: it grows with the store, and it is paid on
every cycle whether the retrieval fires or not.

\section*{Ethics Statement}
\label{pg:eth}
\ours{} maintains a cross-request store of past traffic: for every generated token, the token itself and one hidden-state vector of the serving model. In deployment this store is a cache of user content and must be governed as one, scoped per tenant (as in our evaluation, where each store is built from one benchmark's own requests), retained under the same policy as request logs, and excluded from cross-customer reuse. Hidden states are not encryption, and the text sits verbatim beside them. All experiments use public benchmark data; the one benchmark here that is real user traffic, WildChat, is used as released by its authors, who collected it with user consent.

\bibliography{custom}

\begin{thebibliography}{53}
\providecommand{\natexlab}[1]{#1}

\bibitem[{Alon et~al.(2022)Alon, Xu, He, Sengupta, Roth, and
  Neubig}]{alon2022retomaton}
Uri Alon, Frank~F. Xu, Junxian He, Sudipta Sengupta, Dan Roth, and Graham
  Neubig. 2022.
\newblock \href {https://openreview.net/forum?id=ZJZmKGM6UB} {Neuro-symbolic
  language modeling with automaton-augmented retrieval}.
\newblock In \emph{ICML 2022 Workshop on Knowledge Retrieval and Language
  Models}.

\bibitem[{Barres et~al.(2025)Barres, Dong, Ray, Si, and
  Narasimhan}]{barres2025tau}
Victor Barres, Honghua Dong, Soham Ray, Xujie Si, and Karthik Narasimhan. 2025.
\newblock $\tau^2$-bench: Evaluating conversational agents in a dual-control
  environment.
\newblock \emph{arXiv preprint arXiv:2506.07982}.

\bibitem[{Cai et~al.(2024)Cai, Li, Geng, Peng, Lee, Chen, and
  Dao}]{cai2024medusa}
Tianle Cai, Yuhong Li, Zhengyang Geng, Hongwu Peng, Jason~D. Lee, Deming Chen,
  and Tri Dao. 2024.
\newblock \href {https://proceedings.mlr.press/v235/cai24b.html} {Medusa:
  Simple {LLM} inference acceleration framework with multiple decoding heads}.
\newblock In \emph{Proceedings of the 41st International Conference on Machine
  Learning}, volume 235 of \emph{Proceedings of Machine Learning Research},
  pages 5209--5235. PMLR.

\bibitem[{Chen et~al.(2023)Chen, Borgeaud, Irving, Lespiau, Sifre, and
  Jumper}]{chen2023accelerating}
Charlie Chen, Sebastian Borgeaud, Geoffrey Irving, Jean-Baptiste Lespiau,
  Laurent Sifre, and John Jumper. 2023.
\newblock Accelerating large language model decoding with speculative sampling.
\newblock \emph{arXiv preprint arXiv:2302.01318}.

\bibitem[{Chen et~al.(2021)Chen, Tworek, Jun, Yuan, de~Oliveira~Pinto, Kaplan,
  Edwards, Burda, Joseph, Brockman, Ray, Puri, Krueger, Petrov, Khlaaf, Sastry,
  Mishkin, Chan, Gray, Ryder, Pavlov, Power, Kaiser, Bavarian, Winter, Tillet,
  Such, Cummings, Plappert, Chantzis, Barnes, Herbert-Voss, Guss, Nichol,
  Paino, Tezak, Tang, Babuschkin, Balaji, Jain, Saunders, Hesse, Carr, Leike,
  Achiam, Misra, Morikawa, Radford, Knight, Brundage, Murati, Mayer, Welinder,
  McGrew, Amodei, McCandlish, Sutskever, and Zaremba}]{chen2021codex}
Mark Chen, Jerry Tworek, Heewoo Jun, Qiming Yuan, Henrique~Ponde
  de~Oliveira~Pinto, Jared Kaplan, Harri Edwards, Yuri Burda, Nicholas Joseph,
  Greg Brockman, Alex Ray, Raul Puri, Gretchen Krueger, Michael Petrov, Heidy
  Khlaaf, Girish Sastry, Pamela Mishkin, Brooke Chan, Scott Gray, and 39
  others. 2021.
\newblock \href {https://arxiv.org/abs/2107.03374} {Evaluating large language
  models trained on code}.
\newblock \emph{Preprint}, arXiv:2107.03374.

\bibitem[{Chen et~al.(2026)Chen, Liao, and Wang}]{sense2026}
Shaowen Chen, Zhicheng Liao, and Hongwei Wang. 2026.
\newblock \href {https://arxiv.org/abs/2606.00021} {Sense: Semantic embedding
  navigation with soft-gated evaluation for retrieval-based speculative
  decoding}.
\newblock \emph{Preprint}, arXiv:2606.00021.

\bibitem[{Chen et~al.(2024)Chen, May, Svirschevski, Huang, Ryabinin, Jia, and
  Chen}]{chen2024sequoia}
Zhuoming Chen, Avner May, Ruslan Svirschevski, Yuhsun Huang, Max Ryabinin,
  Zhihao Jia, and Beidi Chen. 2024.
\newblock \href {https://doi.org/10.52202/079017-4116} {Sequoia: Scalable and
  robust speculative decoding}.
\newblock In \emph{Advances in Neural Information Processing Systems},
  volume~37, pages 129531--129563. Curran Associates, Inc.

\bibitem[{Cobbe et~al.(2021)Cobbe, Kosaraju, Bavarian, Chen, Jun, Kaiser,
  Plappert, Tworek, Hilton, Nakano, Hesse, and Schulman}]{cobbe2021gsm8k}
Karl Cobbe, Vineet Kosaraju, Mohammad Bavarian, Mark Chen, Heewoo Jun, Lukasz
  Kaiser, Matthias Plappert, Jerry Tworek, Jacob Hilton, Reiichiro Nakano,
  Christopher Hesse, and John Schulman. 2021.
\newblock \href {https://arxiv.org/abs/2110.14168} {Training verifiers to solve
  math word problems}.
\newblock \emph{Preprint}, arXiv:2110.14168.

\bibitem[{Divilkovskiy et~al.(2025)Divilkovskiy, Malygin, Zlobin, Ilyushin,
  Isali, Kalugin, Aitassova, Yi, and Zeng}]{reader2025}
Maxim Divilkovskiy, Vitaly Malygin, Sergey Zlobin, Stanislav Ilyushin, Sultan
  Isali, Vasily Kalugin, Nuriza Aitassova, Fei Yi, and Weidi Zeng. 2025.
\newblock \href {https://arxiv.org/abs/2508.09072} {Reader: Retrieval-assisted
  drafter for efficient llm inference}.
\newblock \emph{Preprint}, arXiv:2508.09072.

\bibitem[{Dong et~al.(2026)Dong, Wang, Lin, Chen, and
  Hassan}]{semanticspec2026}
Ximing Dong, Shaowei Wang, Dayi Lin, Boyuan Chen, and Ahmed~E. Hassan. 2026.
\newblock \href {https://arxiv.org/abs/2602.03708} {Beyond tokens:
  Semantic-aware speculative decoding for efficient inference by probing
  internal states}.
\newblock \emph{Preprint}, arXiv:2602.03708.

\bibitem[{Drozdov et~al.(2022)Drozdov, Wang, Rahimi, McCallum, Zamani, and
  Iyyer}]{drozdov2022cant}
Andrew Drozdov, Shufan Wang, Razieh Rahimi, Andrew McCallum, Hamed Zamani, and
  Mohit Iyyer. 2022.
\newblock \href {https://doi.org/10.18653/v1/2022.findings-emnlp.218} {You
  can{'}t pick your neighbors, or can you? when and how to rely on retrieval in
  the k{NN}-{LM}}.
\newblock In \emph{Findings of the Association for Computational Linguistics:
  EMNLP 2022}, pages 2997--3007, Abu Dhabi, United Arab Emirates. Association
  for Computational Linguistics.

\bibitem[{Du et~al.(2023)Du, Liu, Wang, Wang, Liu, Chen, Feng, Sha, Peng, and
  Lou}]{du2023classeval}
Xueying Du, Mingwei Liu, Kaixin Wang, Hanlin Wang, Junwei Liu, Yixuan Chen,
  Jiayi Feng, Chaofeng Sha, Xin Peng, and Yiling Lou. 2023.
\newblock \href {https://arxiv.org/abs/2308.01861} {Classeval: A
  manually-crafted benchmark for evaluating llms on class-level code
  generation}.
\newblock \emph{Preprint}, arXiv:2308.01861.

\bibitem[{Fang et~al.(2025)Fang, Fu, Zhao, and Wang}]{respec2025}
Min Fang, Zhihui Fu, Qibin Zhao, and Jun Wang. 2025.
\newblock \href {https://arxiv.org/abs/2511.01282} {When, what, and how:
  Rethinking retrieval-enhanced speculative decoding}.
\newblock \emph{Preprint}, arXiv:2511.01282.

\bibitem[{Fu et~al.(2024)Fu, Bailis, Stoica, and Zhang}]{fu2024lookahead}
Yichao Fu, Peter Bailis, Ion Stoica, and Hao Zhang. 2024.
\newblock \href {https://proceedings.mlr.press/v235/fu24a.html} {Break the
  sequential dependency of {LLM} inference using lookahead decoding}.
\newblock In \emph{Proceedings of the 41st International Conference on Machine
  Learning}, volume 235 of \emph{Proceedings of Machine Learning Research},
  pages 14060--14079. PMLR.

\bibitem[{Gritta et~al.(2025)Gritta, Xue, and Lampouras}]{dresd2025}
Milan Gritta, Huiyin Xue, and Gerasimos Lampouras. 2025.
\newblock \href {https://doi.org/10.18653/v1/2025.findings-acl.1017}
  {{DR}e{SD}: Dense retrieval for speculative decoding}.
\newblock In \emph{Findings of the Association for Computational Linguistics:
  ACL 2025}, pages 19822--19832, Vienna, Austria. Association for Computational
  Linguistics.

\bibitem[{He et~al.(2021)He, Neubig, and Berg-Kirkpatrick}]{he2021efficient}
Junxian He, Graham Neubig, and Taylor Berg-Kirkpatrick. 2021.
\newblock \href {https://doi.org/10.18653/v1/2021.emnlp-main.461} {Efficient
  nearest neighbor language models}.
\newblock In \emph{Proceedings of the 2021 Conference on Empirical Methods in
  Natural Language Processing}, pages 5703--5714, Online and Punta Cana,
  Dominican Republic. Association for Computational Linguistics.

\bibitem[{He et~al.(2024)He, Zhong, Cai, Lee, and He}]{he2024rest}
Zhenyu He, Zexuan Zhong, Tianle Cai, Jason Lee, and Di~He. 2024.
\newblock \href {https://doi.org/10.18653/v1/2024.naacl-long.88} {{REST}:
  Retrieval-based speculative decoding}.
\newblock In \emph{Proceedings of the 2024 Conference of the North American
  Chapter of the Association for Computational Linguistics: Human Language
  Technologies (Volume 1: Long Papers)}, pages 1582--1595, Mexico City, Mexico.
  Association for Computational Linguistics.

\bibitem[{Ho et~al.(2024)Ho, Park, and Wang}]{crest2024}
Sophia Ho, Jinsol Park, and Patrick Wang. 2024.
\newblock \href {https://arxiv.org/abs/2408.04678} {Crest: Effectively
  compacting a datastore for retrieval-based speculative decoding}.
\newblock \emph{Preprint}, arXiv:2408.04678.

\bibitem[{Hu et~al.(2025)Hu, Wang, Zhang, Zhang, Li, Chen, and
  Zhang}]{hu2025samdecoding}
Yuxuan Hu, Ke~Wang, Xiaokang Zhang, Fanjin Zhang, Cuiping Li, Hong Chen, and
  Jing Zhang. 2025.
\newblock \href {https://doi.org/10.18653/v1/2025.acl-long.595} {{SAM}
  decoding: Speculative decoding via suffix automaton}.
\newblock In \emph{Proceedings of the 63rd Annual Meeting of the Association
  for Computational Linguistics (Volume 1: Long Papers)}, pages 12187--12204,
  Vienna, Austria. Association for Computational Linguistics.

\bibitem[{Jimenez et~al.(2024)Jimenez, Yang, Wettig, Yao, Pei, Press, and
  Narasimhan}]{jimenez2024swebench}
Carlos~E Jimenez, John Yang, Alexander Wettig, Shunyu Yao, Kexin Pei, Ofir
  Press, and Karthik~R Narasimhan. 2024.
\newblock \href {https://openreview.net/forum?id=VTF8yNQM66} {{SWE}-bench: Can
  language models resolve real-world github issues?}
\newblock In \emph{The Twelfth International Conference on Learning
  Representations}.

\bibitem[{Jin et~al.(2026)Jin, Nguyen, and Inoue}]{goose2026}
Tao Jin, Phuong~Minh Nguyen, and Naoya Inoue. 2026.
\newblock \href {https://arxiv.org/abs/2604.02047} {Goose: Anisotropic
  speculation trees for training-free speculative decoding}.
\newblock \emph{Preprint}, arXiv:2604.02047.
\newblock To appear at the Conference on Language Modeling (COLM).

\bibitem[{Khandelwal et~al.(2020)Khandelwal, Levy, Jurafsky, Zettlemoyer, and
  Lewis}]{khandelwal2020knnlm}
Urvashi Khandelwal, Omer Levy, Dan Jurafsky, Luke Zettlemoyer, and Mike Lewis.
  2020.
\newblock \href {https://openreview.net/forum?id=HklBjCEKvH} {Generalization
  through memorization: Nearest neighbor language models}.
\newblock In \emph{International Conference on Learning Representations}.

\bibitem[{Leviathan et~al.(2023)Leviathan, Kalman, and
  Matias}]{leviathan2023fast}
Yaniv Leviathan, Matan Kalman, and Yossi Matias. 2023.
\newblock \href {https://proceedings.mlr.press/v202/leviathan23a.html} {Fast
  inference from transformers via speculative decoding}.
\newblock In \emph{Proceedings of the 40th International Conference on Machine
  Learning}, volume 202 of \emph{Proceedings of Machine Learning Research},
  pages 19274--19286. PMLR.

\bibitem[{Li et~al.(2021)Li, Arora, Chen, Gupta, Gupta, and
  Mehdad}]{li2021mtop}
Haoran Li, Abhinav Arora, Shuohui Chen, Anchit Gupta, Sonal Gupta, and Yashar
  Mehdad. 2021.
\newblock \href {https://doi.org/10.18653/v1/2021.eacl-main.257} {{MTOP}: A
  comprehensive multilingual task-oriented semantic parsing benchmark}.
\newblock In \emph{Proceedings of the 16th Conference of the European Chapter
  of the Association for Computational Linguistics: Main Volume}, pages
  2950--2962, Online. Association for Computational Linguistics.

\bibitem[{Li et~al.(2026)Li, Xu, Li, Yang, Xu, Yin, Li, Ngai, and
  Barsoum}]{loosesd2026}
Jinze Li, Yixing Xu, Guanchen Li, Shuo Yang, Jinfeng Xu, Xuanwu Yin, Dong Li,
  Edith C.~H. Ngai, and Emad Barsoum. 2026.
\newblock \href {https://openreview.net/forum?id=JjoTg34YiU} {Training-free
  loosely speculative decoding: Accepting semantically correct drafts beyond
  exact match}.
\newblock In \emph{The Fourteenth International Conference on Learning
  Representations}.

\bibitem[{Li et~al.(2024{\natexlab{a}})Li, Chen, Holtzman, Chen, Lin, Yih, and
  Lin}]{li2024nearest}
Minghan Li, Xilun Chen, Ari Holtzman, Beidi Chen, Jimmy Lin, Wen-tau Yih, and
  Xi~Victoria Lin. 2024{\natexlab{a}}.
\newblock \href {https://doi.org/10.52202/079017-2574} {Nearest neighbor
  speculative decoding for llm generation and attribution}.
\newblock In \emph{Advances in Neural Information Processing Systems},
  volume~37, pages 80987--81015. Curran Associates, Inc.

\bibitem[{Li et~al.(2023)Li, Zhao, Yu, Song, Li, Yu, Li, Huang, and
  Li}]{li2023apibank}
Minghao Li, Yingxiu Zhao, Bowen Yu, Feifan Song, Hangyu Li, Haiyang Yu, Zhoujun
  Li, Fei Huang, and Yongbin Li. 2023.
\newblock \href {https://doi.org/10.18653/v1/2023.emnlp-main.187} {{API}-bank:
  A comprehensive benchmark for tool-augmented {LLM}s}.
\newblock In \emph{Proceedings of the 2023 Conference on Empirical Methods in
  Natural Language Processing}, pages 3102--3116, Singapore. Association for
  Computational Linguistics.

\bibitem[{Li et~al.(2024{\natexlab{b}})Li, Wei, Zhang, and
  Zhang}]{li2024eagle2}
Yuhui Li, Fangyun Wei, Chao Zhang, and Hongyang Zhang. 2024{\natexlab{b}}.
\newblock \href {https://doi.org/10.18653/v1/2024.emnlp-main.422} {{EAGLE}-2:
  Faster inference of language models with dynamic draft trees}.
\newblock In \emph{Proceedings of the 2024 Conference on Empirical Methods in
  Natural Language Processing}, pages 7421--7432, Miami, Florida, USA.
  Association for Computational Linguistics.

\bibitem[{Li et~al.(2024{\natexlab{c}})Li, Wei, Zhang, and Zhang}]{li2024eagle}
Yuhui Li, Fangyun Wei, Chao Zhang, and Hongyang Zhang. 2024{\natexlab{c}}.
\newblock \href {https://proceedings.mlr.press/v235/li24bt.html} {{EAGLE}:
  Speculative sampling requires rethinking feature uncertainty}.
\newblock In \emph{Proceedings of the 41st International Conference on Machine
  Learning}, volume 235 of \emph{Proceedings of Machine Learning Research},
  pages 28935--28948. PMLR.

\bibitem[{Li et~al.(2025)Li, Wei, Zhang, and Zhang}]{li2025eagle3}
Yuhui Li, Fangyun Wei, Chao Zhang, and Hongyang Zhang. 2025.
\newblock \href {https://openreview.net/forum?id=4exx1hUffq} {{EAGLE}-3:
  Scaling up inference acceleration of large language models via training-time
  test}.
\newblock In \emph{The Thirty-ninth Annual Conference on Neural Information
  Processing Systems}.

\bibitem[{Lin et~al.(2026)Lin, Zhen, Yang, Wang, Liu, Chen, Zhang, Zhou, Li,
  Chen, Li, Yang, Li, Yu, Dong, Yuan, and Wang}]{agentinfer2025}
Weizhe Lin, Hui-Ling Zhen, Shuai Yang, Xian Wang, Renxi Liu, Hanting Chen,
  Wangze Zhang, Chuansai Zhou, Yiming Li, Chen Chen, Xing Li, Zhiyuan Yang,
  Xiaosong Li, Xianzhi Yu, Zhenhua Dong, Mingxuan Yuan, and Yunhe Wang. 2026.
\newblock \href {https://arxiv.org/abs/2512.18337} {Towards efficient agents: A
  co-design of inference architecture and system}.
\newblock \emph{Preprint}, arXiv:2512.18337.

\bibitem[{Liu et~al.(2026{\natexlab{a}})Liu, Xie, Hu, Xiao, and
  Huang}]{adapld2026}
Runheng Liu, Jincheng Xie, Wen Hu, Xingchen Xiao, and Heyan Huang.
  2026{\natexlab{a}}.
\newblock \href {https://arxiv.org/abs/2606.05742} {Adapld: Adaptive retrieval
  and reuse for efficient model-free speculative decoding}.
\newblock \emph{Preprint}, arXiv:2606.05742.

\bibitem[{Liu et~al.(2026{\natexlab{b}})Liu, Lv, Li, Gao, Sun, and
  Sun}]{logitspec2026}
Tianyu Liu, Qitan Lv, Hao Li, Xing Gao, Xiao Sun, and Xiaoyan Sun.
  2026{\natexlab{b}}.
\newblock \href {https://openreview.net/forum?id=I9f5NdjBvK} {Logitspec:
  Accelerating retrieval-based speculative decoding via next next token
  speculation}.

\bibitem[{Luo et~al.(2025)Luo, Wang, Zhu, Zhang, Zhang, Yang, and
  Xu}]{luo2024recycling}
Xianzhen Luo, Yixuan Wang, Qingfu Zhu, Zhiming Zhang, Xuanyu Zhang, Qing Yang,
  and Dongliang Xu. 2025.
\newblock \href {https://doi.org/10.18653/v1/2025.acl-long.338} {Turning trash
  into treasure: Accelerating inference of large language models with token
  recycling}.
\newblock In \emph{Proceedings of the 63rd Annual Meeting of the Association
  for Computational Linguistics (Volume 1: Long Papers)}, pages 6816--6831,
  Vienna, Austria. Association for Computational Linguistics.

\bibitem[{Miao et~al.(2024)Miao, Oliaro, Zhang, Cheng, Wang, Zhang, Wong, Zhu,
  Yang, Shi, Shi, Chen, Arfeen, Abhyankar, and Jia}]{miao2024specinfer}
Xupeng Miao, Gabriele Oliaro, Zhihao Zhang, Xinhao Cheng, Zeyu Wang, Zhengxin
  Zhang, Rae Ying~Yee Wong, Alan Zhu, Lijie Yang, Xiaoxiang Shi, Chunan Shi,
  Zhuoming Chen, Daiyaan Arfeen, Reyna Abhyankar, and Zhihao Jia. 2024.
\newblock \href {https://doi.org/10.1145/3620666.3651335} {Specinfer:
  Accelerating large language model serving with tree-based speculative
  inference and verification}.
\newblock In \emph{Proceedings of the 29th ACM International Conference on
  Architectural Support for Programming Languages and Operating Systems, Volume
  3}, ASPLOS '24, page 932–949, New York, NY, USA. Association for Computing
  Machinery.

\bibitem[{Oliaro et~al.(2025)Oliaro, Jia, Campos, and
  Qiao}]{oliaro2024suffixdecoding}
Gabriele Oliaro, Zhihao Jia, Daniel~F Campos, and Aurick Qiao. 2025.
\newblock \href {https://openreview.net/forum?id=uwL0vbeEVn} {Suffixdecoding:
  Extreme speculative decoding for emerging {AI} applications}.
\newblock In \emph{The Thirty-ninth Annual Conference on Neural Information
  Processing Systems}.

\bibitem[{Quan et~al.(2025)Quan, Feng, Hao, Jiang, Zhang, and
  Wang}]{quan2025rasd}
Guofeng Quan, Wenfeng Feng, Chuzhan Hao, Guochao Jiang, Yuewei Zhang, and
  Hao~Henry Wang. 2025.
\newblock \href {https://doi.org/10.18653/v1/2025.findings-acl.320} {{RASD}:
  Retrieval-augmented speculative decoding}.
\newblock In \emph{Findings of the Association for Computational Linguistics:
  ACL 2025}, pages 6167--6177, Vienna, Austria. Association for Computational
  Linguistics.

\bibitem[{Saxena(2023)}]{saxena2023pld}
Apoorv Saxena. 2023.
\newblock Prompt lookup decoding.
\newblock \url{https://github.com/apoorvumang/prompt-lookup-decoding}.

\bibitem[{Shen et~al.(2026)Shen, Liu, Hu, Kong, Zhang, Dai, Zhang, Ge, Chen,
  Li, Wan, and Wang}]{graft2026}
Yuhao Shen, Tianyu Liu, Xinyi Hu, Quan Kong, Baolin Zhang, Jun Dai, Jun Zhang,
  Shuang Ge, Lei Chen, Yue Li, Mingcheng Wan, and Cong Wang. 2026.
\newblock \href {https://arxiv.org/abs/2605.20104} {Draft less, retrieve more:
  Hybrid tree construction for speculative decoding}.
\newblock \emph{Preprint}, arXiv:2605.20104.

\bibitem[{Suzgun et~al.(2023)Suzgun, Scales, Sch{\"a}rli, Gehrmann, Tay, Chung,
  Chowdhery, Le, Chi, Zhou, and Wei}]{suzgun2023bbh}
Mirac Suzgun, Nathan Scales, Nathanael Sch{\"a}rli, Sebastian Gehrmann, Yi~Tay,
  Hyung~Won Chung, Aakanksha Chowdhery, Quoc~V. Le, Ed~H. Chi, Denny Zhou, and
  Jason Wei. 2023.
\newblock Challenging {BIG}-bench tasks and whether chain-of-thought can solve
  them.
\newblock In \emph{Findings of ACL}.

\bibitem[{Tang et~al.(2023)Tang, Deng, Lin, Han, Liang, Cao, and
  Sun}]{tang2023toolalpaca}
Qiaoyu Tang, Ziliang Deng, Hongyu Lin, Xianpei Han, Qiao Liang, Boxi Cao, and
  Le~Sun. 2023.
\newblock \href {https://arxiv.org/abs/2306.05301} {Toolalpaca: Generalized
  tool learning for language models with 3000 simulated cases}.
\newblock \emph{Preprint}, arXiv:2306.05301.

\bibitem[{Valmeekam et~al.(2023)Valmeekam, Marquez, Olmo, Sreedharan, and
  Kambhampati}]{valmeekam2023planbench}
Karthik Valmeekam, Matthew Marquez, Alberto Olmo, Sarath Sreedharan, and
  Subbarao Kambhampati. 2023.
\newblock \href {https://doi.org/10.52202/075280-1693} {Planbench: An
  extensible benchmark for evaluating large language models on planning and
  reasoning about change}.
\newblock In \emph{Advances in Neural Information Processing Systems},
  volume~36, pages 38975--38987. Curran Associates, Inc.

\bibitem[{Wang et~al.(2023)Wang, Song, Drozdov, Garimella, Manjunatha, and
  Iyyer}]{wang2023knnlm}
Shufan Wang, Yixiao Song, Andrew Drozdov, Aparna Garimella, Varun Manjunatha,
  and Mohit Iyyer. 2023.
\newblock \href {https://openreview.net/forum?id=3FNrGv5MKb} {\$k\${NN}-{LM}
  does not improve open-ended text generation}.
\newblock In \emph{The 2023 Conference on Empirical Methods in Natural Language
  Processing}.

\bibitem[{Wei et~al.(2024)Wei, Wang, Liu, Ding, and Zhang}]{wei2024magicoder}
Yuxiang Wei, Zhe Wang, Jiawei Liu, Yifeng Ding, and Lingming Zhang. 2024.
\newblock \href {https://proceedings.mlr.press/v235/wei24h.html} {Magicoder:
  Empowering code generation with {OSS}-instruct}.
\newblock In \emph{Proceedings of the 41st International Conference on Machine
  Learning}, volume 235 of \emph{Proceedings of Machine Learning Research},
  pages 52632--52657. PMLR.

\bibitem[{Xia et~al.(2026{\natexlab{a}})Xia, Ribar, and
  Balanca}]{relaxedsd2026}
Guoxuan Xia, Luka Ribar, and Paul Balanca. 2026{\natexlab{a}}.
\newblock \href {https://arxiv.org/abs/2607.08690} {A practical investigation
  of training-free relaxed speculative decoding}.
\newblock \emph{Preprint}, arXiv:2607.08690.

\bibitem[{Xia et~al.(2026{\natexlab{b}})Xia, Li, Du, Song, and
  Li}]{toolspec2026}
Heming Xia, Yongqi Li, Cunxiao Du, Mingbo Song, and Wenjie Li.
  2026{\natexlab{b}}.
\newblock \href {https://arxiv.org/abs/2604.13519} {Toolspec: Accelerating tool
  calling via schema-aware and retrieval-augmented speculative decoding}.
\newblock \emph{Preprint}, arXiv:2604.13519.

\bibitem[{Xia et~al.(2024)Xia, Yang, Dong, Wang, Li, Ge, Liu, Li, and
  Sui}]{xia2024specbench}
Heming Xia, Zhe Yang, Qingxiu Dong, Peiyi Wang, Yongqi Li, Tao Ge, Tianyu Liu,
  Wenjie Li, and Zhifang Sui. 2024.
\newblock \href {https://doi.org/10.18653/v1/2024.findings-acl.456} {Unlocking
  efficiency in large language model inference: A comprehensive survey of
  speculative decoding}.
\newblock In \emph{Findings of the Association for Computational Linguistics:
  ACL 2024}, pages 7655--7671, Bangkok, Thailand. Association for Computational
  Linguistics.

\bibitem[{Yang et~al.(2023)Yang, Ge, Wang, Jiao, Jiang, Yang, Majumder, and
  Wei}]{yang2023llma}
Nan Yang, Tao Ge, Liang Wang, Binxing Jiao, Daxin Jiang, Linjun Yang, Rangan
  Majumder, and Furu Wei. 2023.
\newblock \href {https://arxiv.org/abs/2304.04487} {Inference with reference:
  Lossless acceleration of large language models}.
\newblock \emph{Preprint}, arXiv:2304.04487.

\bibitem[{Yao et~al.(2025)Yao, Shinn, Razavi, and Narasimhan}]{yao2024tau}
Shunyu Yao, Noah Shinn, Pedram Razavi, and Karthik~R Narasimhan. 2025.
\newblock \href {https://openreview.net/forum?id=roNSXZpUDN}
  {\{\${\textbackslash}tau\$\}-bench: A benchmark for
  {\textbackslash}underline\{T\}ool-{\textbackslash}underline\{A\}gent-{\textbackslash}underline\{U\}ser
  interaction in real-world domains}.
\newblock In \emph{The Thirteenth International Conference on Learning
  Representations}.

\bibitem[{Yu et~al.(2018)Yu, Zhang, Yang, Yasunaga, Wang, Li, Ma, Li, Yao,
  Roman, Zhang, and Radev}]{yu2018spider}
Tao Yu, Rui Zhang, Kai Yang, Michihiro Yasunaga, Dongxu Wang, Zifan Li, James
  Ma, Irene Li, Qingning Yao, Shanelle Roman, Zilin Zhang, and Dragomir Radev.
  2018.
\newblock \href {https://doi.org/10.18653/v1/D18-1425} {{S}pider: A large-scale
  human-labeled dataset for complex and cross-domain semantic parsing and
  text-to-{SQL} task}.
\newblock In \emph{Proceedings of the 2018 Conference on Empirical Methods in
  Natural Language Processing}, pages 3911--3921, Brussels, Belgium.
  Association for Computational Linguistics.

\bibitem[{Zhang et~al.(2026)Zhang, Li, Zhang, and Zhao}]{racer2025}
Zihong Zhang, Zuchao Li, Lefei Zhang, and Hai Zhao. 2026.
\newblock \href {https://openreview.net/forum?id=i8SjqQQOVt} {{RACER}:
  Retrieval-augmented contextual rapid speculative decoding}.

\bibitem[{Zhao et~al.(2024)Zhao, Ren, Hessel, Cardie, Choi, and
  Deng}]{zhao2024wildchat}
Wenting Zhao, Xiang Ren, Jack Hessel, Claire Cardie, Yejin Choi, and Yuntian
  Deng. 2024.
\newblock \href {https://openreview.net/forum?id=Bl8u7ZRlbM} {Wildchat: 1m
  chat{GPT} interaction logs in the wild}.
\newblock In \emph{The Twelfth International Conference on Learning
  Representations}.

\bibitem[{Zheng et~al.(2023)Zheng, Chiang, Sheng, Zhuang, Wu, Zhuang, Lin, Li,
  Li, Xing, Zhang, Gonzalez, and Stoica}]{zheng2023mtbench}
Lianmin Zheng, Wei-Lin Chiang, Ying Sheng, Siyuan Zhuang, Zhanghao Wu, Yonghao
  Zhuang, Zi~Lin, Zhuohan Li, Dacheng Li, Eric Xing, Hao Zhang, Joseph
  Gonzalez, and Ion Stoica. 2023.
\newblock \href {https://doi.org/10.52202/075280-2020} {Judging llm-as-a-judge
  with mt-bench and chatbot arena}.
\newblock In \emph{Advances in Neural Information Processing Systems},
  volume~36, pages 46595--46623. Curran Associates, Inc.

\end{thebibliography}

\clearpage
\appendix

\section{Experimental Details}
\label{app:matrix}\label{app:repro}\label{app:hardware}\label{app:hyper}\label{app:baselines}

\begin{table}[H]\centering\small
\setlength{\tabcolsep}{4pt}
\begin{tabular}{@{}l@{\;\;}r@{\;\;}r@{}}
\toprule
\multicolumn{3}{@{}l}{\emph{(a) benchmarks}} \\
\textbf{Benchmark} & \textbf{Requests} & \textbf{Max tokens} \\
\midrule
API-Bank & 597 & 1{,}024 \\
ToolAlpaca & 100 & 1{,}024 \\
tau-bench ret.\ & 69 & 1{,}024 \\
HumanEval & 164 & \phantom{0,0}512 \\
ClassEval & 100 & 2{,}048 \\
GSM8K & 1{,}319 & 1{,}024 \\
MT-Bench & 80 & 1{,}024 \\
WildChat & 200 & 2{,}048 \\
\midrule
\multicolumn{3}{@{}l}{\emph{(b) hardware and software}} \\
\midrule
GPU, 8B and 14B & \multicolumn{2}{r@{}}{NVIDIA A40, 48\,GB} \\
GPU, 32B & \multicolumn{2}{r@{}}{NVIDIA H100, 80\,GB} \\
GPU, ablations & \multicolumn{2}{r@{}}{NVIDIA RTX~3090, 24\,GB} \\
CUDA / PyTorch & \multicolumn{2}{r@{}}{12.1 / 2.4.1} \\
Precision, batch & \multicolumn{2}{r@{}}{bfloat16, 1} \\
\bottomrule
\end{tabular}
\caption{\textbf{Benchmarks, generation caps, and the hardware and software configuration.} \emph{(a)} requests each benchmark contributes and the
largest number of tokens one request may generate before we stop it. Caps are inherited from the
system we build on, except ClassEval: there the inherited 512 truncated 98 of 100 generations for
every method, so all arms were re-run at 2{,}048. Only WildChat is subsampled, having no evaluation
split. \emph{(b)} one frozen configuration (Algorithm~\ref{alg:step}) produces every number here, on
one accelerator per run: the A40 the 8B and 14B cells, the H100 the 32B block, the RTX~3090 the
ablations. Accepted length transfers between these cards; wall-clock does not. Every arm is compared
token for token against an independent greedy run: the twelve arms match on 91--100\% of samples (median
97\%), the residue being bfloat16 tie-breaking; \ours{} and GOOSE differ by at most 4.3
points on any cell.}
\label{tab:matrix}\label{tab:repro}\label{tab:hardware}
\end{table}

\section{Additional Ablations and Analysis}
\label{app:rootonly}\label{app:tables}\label{app:batch}

\begin{table}[H]\centering\small
\setlength{\tabcolsep}{3.5pt}\renewcommand{\arraystretch}{0.94}
\begin{tabular}{@{}l r r r r@{}}
\toprule
\textbf{Draft source} & \textbf{alone} & \textbf{$+$key} & \textbf{additive} & \textbf{$\Delta$} \\
\midrule
token adjacency  & 2.94 & 5.87 & 4.11 & $+$43\% \\
suffix automaton & 5.34 & 5.94 & 6.51 & $-$9\% \\
both (deployed)  & 6.36 & 6.75 & 7.52 & $-$10\% \\
\bottomrule
\end{tabular}
\caption{\textbf{Accepted length for each draft source alone and with the semantic key merged in.} Measured on
API-Bank, Llama-3.1-8B, entire benchmark ($n{=}597$, two allocations).
\emph{alone}: that source drafting without the semantic key. \emph{$+$key}: the same source with the
key merged in. \emph{additive}: what the pair would reach if the two never overlapped---the source
alone plus the key's own increment over an empty tree ($+$1.17 $\tau$, 2.76 to 3.92).
\emph{$\Delta$}: the measured pair against that expectation, so a positive value means the two do
more together than apart and a negative one that they cover the same positions. The same key lands
43\% above additive on the host that supplies grafting breadth but no depth, and below it on the
host that already supplies depth.}
\label{tab:interaction}
\end{table}

\begin{table}[H]\centering\small
\setlength{\tabcolsep}{3.5pt}\renewcommand{\arraystretch}{0.94}
\begin{tabular}{@{}l r r@{}}
\toprule
\textbf{Method (its key)} & \textbf{reaches} & \textbf{only it} \\
\midrule
exact suffix (PLD, SuffixDec., SAM) & 92.5 & 0.2 \\
token adjacency (Token Recycling) & 86.7 & 0.4 \\
static token embedding & 90.2 & 0.2 \\
per-position hidden state (ours) & \textbf{96.5} & \textbf{3.3} \\
\midrule
suffix $\cup$ adjacency (both lexical) & 95.6 & \\
\quad $\cup$ ours (all three) & 99.0 & \\
\bottomrule
\end{tabular}
\caption{\textbf{Coverage of the recoverable positions, key by key.} How much of the reachable past
each training-free key can actually address, on API-Bank with Llama-3.1-8B. The universe---9{,}244
of the same 10{,}313 API-Bank positions---is those whose two-token future occurs verbatim in a
strictly-past trace, so a copying drafter could in principle reach them. \emph{reaches}: the share
of that universe at which the key puts the true next token among its eight candidates.
\emph{only it}: the share no other key in the table reaches. Ours is near-independent of both
lexical keys (lift 1.013 and 1.009, against 1.042 between them) and recovers 81\% and 91\% of what
they respectively miss. Reaching a position is not delivering it.}
\label{tab:venn}
\end{table}

A batched replica of the tree verifier (masked greedy over left-padded rows) runs Llama-3.1-8B on
API-Bank and GSM8K at batch sizes 1--32, twelve arms from 2-deep chains to 32-node trees, each
priced against autoregressive decoding at the same batch size over the same rows. It is a separate
harness, comparable only within itself. The semantic key is queried per position here, as in
deployment; what it lacks is the similarity-sized depth schedule, so it copies fixed-length spans.

%
%
\begin{figure}[t]\centering
\begin{tikzpicture}
\begin{axis}[width=3.05cm, height=3.45cm, scale only axis, font=\scriptsize,
    axis line style={black!55}, tick align=outside, tick style={black!55},
    xmajorgrids, ymajorgrids, grid style={black!12, line width=0.35pt}, axis on top,
    xmode=log, log basis x=2, xtick={1,4,8,16,32}, xticklabels={1,4,8,16,32},
    xmin=0.85, xmax=38, ymin=0.33, ymax=1.68, ytick={0.5,1.0,1.5},
    xticklabel style={font=\tiny}, yticklabel style={font=\tiny},
    title style={font=\scriptsize, yshift=-3pt}, name=A, title={API-Bank},
    ylabel={tok/s over AR at the same $B$}, ylabel style={font=\scriptsize, yshift=-4pt}]
  \addplot[black!45, densely dashed, line width=0.5pt, forget plot] coordinates {(0.85,1)(38,1)};
  \addplot[gkGray, dotted, thick, mark=*, mark size=1.3pt] coordinates {(1,1.28) (4,1.19) (8,1.14) (16,1.10) (32,1.08)};
  \addplot[gkGray, densely dotted, thick, mark=square*, mark size=1.3pt] coordinates {(1,1.34) (4,1.22) (8,1.17) (16,1.12) (32,1.08)};
  \addplot[gkGray, dashed, thick, mark=triangle*, mark size=1.3pt] coordinates {(1,1.40) (4,1.23) (8,1.16) (16,1.09) (32,0.96)};
  \addplot[cbBlue, solid, thick, mark=*, mark size=1.3pt] coordinates {(1,1.53) (4,1.46) (8,1.34) (16,1.29) (32,1.11)};
  \addplot[gkOrange, solid, thick, mark=square*, mark size=1.3pt] coordinates {(1,1.59) (4,1.43) (8,1.34) (16,1.13) (32,0.98)};
  \addplot[cbRed, solid, thick, mark=triangle*, mark size=1.3pt] coordinates {(1,1.53) (4,1.41) (8,1.16) (16,0.96) (32,0.77)};
\end{axis}
\begin{axis}[width=3.05cm, height=3.45cm, scale only axis, font=\scriptsize,
    axis line style={black!55}, tick align=outside, tick style={black!55},
    xmajorgrids, ymajorgrids, grid style={black!12, line width=0.35pt}, axis on top,
    xmode=log, log basis x=2, xtick={1,4,8,16,32}, xticklabels={1,4,8,16,32},
    xmin=0.85, xmax=38, ymin=0.33, ymax=1.68, ytick={0.5,1.0,1.5},
    xticklabel style={font=\tiny}, yticklabel style={font=\tiny},
    title style={font=\scriptsize, yshift=-3pt}, at={([xshift=0.95cm]A.south east)}, anchor=south west,
    title={GSM8K}, yticklabels={,,}]
  \addplot[black!45, densely dashed, line width=0.5pt, forget plot] coordinates {(0.85,1)(38,1)};
  \addplot[gkGray, dotted, thick, mark=*, mark size=1.3pt] coordinates {(1,1.10) (4,1.01) (8,0.98) (16,0.93) (32,0.95)};
  \addplot[gkGray, densely dotted, thick, mark=square*, mark size=1.3pt] coordinates {(1,1.11) (4,0.99) (8,0.93) (16,0.91) (32,0.88)};
  \addplot[gkGray, dashed, thick, mark=triangle*, mark size=1.3pt] coordinates {(1,1.11) (4,0.98) (8,0.90) (16,0.87) (32,0.65)};
  \addplot[cbBlue, solid, thick, mark=*, mark size=1.3pt] coordinates {(1,1.27) (4,1.12) (8,1.03) (16,1.00) (32,0.82)};
  \addplot[gkOrange, solid, thick, mark=square*, mark size=1.3pt] coordinates {(1,1.27) (4,1.08) (8,1.02) (16,0.76) (32,0.62)};
  \addplot[cbRed, solid, thick, mark=triangle*, mark size=1.3pt] coordinates {(1,1.46) (4,1.15) (8,0.85) (16,0.58) (32,0.41)};
\end{axis}
\node[anchor=north, font=\scriptsize] at ([yshift=-11pt, xshift=0.475cm]A.south east) {batch size $B$};
\node[anchor=north, font=\tiny, align=center] at ([yshift=-21pt, xshift=0.475cm]A.south east)
  {\textcolor{gkGray}{\rule{4pt}{4pt}}~chains (2/4/8)\quad
   \textcolor{cbBlue}{\rule{4pt}{4pt}}~tree-8\quad
   \textcolor{gkOrange}{\rule{4pt}{4pt}}~tree-16\quad
   \textcolor{cbRed}{\rule{4pt}{4pt}}~tree-32};
\end{tikzpicture}
\caption{\textbf{The same experiment run at larger batch sizes.} Throughput over autoregressive
decoding at the same batch size, as the batch grows from 1 to 32; the dashed line is parity, and an
arm below it is slower than not speculating at all. Wider trees drop through parity sooner:
$B^{*}$, the largest measured batch at which an arm still reaches $1\times$, is written below as
\emph{arm}~\emph{API-Bank}/\emph{GSM8K}: chain-2~32/4, chain-4~32/1, chain-8~16/1, tree-8~32/16,
tree-16~16/8, tree-32~8/4. A chain-$n$ arm drafts one chain $n$ tokens deep; a tree-$n$ arm drafts a
tree of $n$ nodes. The semantic key
is queried per position here, as in deployment, but copies fixed-length spans rather than depths set
by the match's own similarity; it moves accepted length by $-$2.4 to $+$2.7\%, inside this harness's
repeat noise, and buys more nodes than it earns tokens in 19 of the 20 cells. Llama-3.1-8B; the store is
built from 96 earlier requests and then held fixed, so every arm sees the same pool; three trials,
64 generated tokens per request.}
\label{fig:batchprobe}
\end{figure}
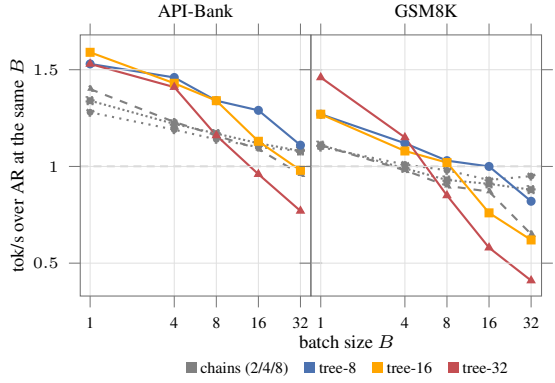

\begin{table}[t]\centering\small
\setlength{\tabcolsep}{3.5pt}\renewcommand{\arraystretch}{0.94}
\begin{tabular}{@{}p{6.55cm} r@{}}
\toprule
\multicolumn{2}{@{}l}{\emph{(a) draft sources, taken away one at a time}} \\
\midrule
\textbf{Configuration} & \boldmath$\tau$ \\
\midrule
merged system & 6.16 \\
\;lexical-only (semantic entry budget zeroed) & 5.95 \\
\;semantic only (both lexical sources off; root branching declared) & 3.75 \\
\;\;\;\emph{null control}: the same arm retrieving nothing & 2.67 \\
\midrule
merged system & 4.86 \\
\;one source chosen per step instead of merged (SAM-Decoding's mechanism) & 3.51 \\
\bottomrule
\end{tabular}

\vspace{2pt}
\begin{tabular}{@{}l r@{\;}r@{\;}r@{}}
\toprule
\multicolumn{4}{@{}l}{\emph{(b) the merge itself, taken away}} \\
\midrule
 & Merged & Root-attached & Lexical-only \\
\midrule
API-Bank $\tau$ & 5.246 & 4.801 & 5.042 \\
tau-bench retail $\tau$ & 3.649 & 3.396 & 3.434 \\
GSM8K $\tau$ & 2.961 & 2.928 & 2.995 \\
\bottomrule
\end{tabular}
\caption{\textbf{Draft sources, and then the merge itself, removed one at a time.} \emph{(a)} one configuration per
row: the deployed system; the same system with the semantic source's budget zeroed; with both
lexical sources off instead; and a control that keeps that arm but retrieves nothing. Below the
rule, the deployed system against choosing one source per step rather than merging them. The
semantic source alone loses to the lexical-only arm, so the merge is not a wrapper around a
stronger drafter. \emph{(b)} the merge itself removed: \emph{root-attached} hangs every retrieved
chain from the root as its own branch---the usual multi-source arrangement---at an unchanged node
budget, and that drives the source below the lexical-only arm on both repetitive workloads.
Nodes differ as sharply as sources do: crediting each node to the source that placed it, 55\% of
the automaton's longest-match chain is accepted, against 5\% of its frequency-ranked branches and
3\% of token adjacency.
Llama-3.1-8B, two repeats throughout; (a) first block $n{=}200$, second $n{=}100$, (b) $n{=}50$. Both
lexical sources are verifiably absent from the semantic arms, by per-node counters rather than by
environment strings.}
\label{tab:arms}
\end{table}

\begin{table*}[!ht]\centering
\caption{\textbf{The five non-repetitive workloads.} The five methods of
Table~\ref{tab:main-echo}, on the workloads where \S\ref{sec:gap}'s diagnosis says the semantic
source has little to find. Protocol of Table~\ref{tab:main-echo}: entire benchmark, greedy, one
machine for the whole table, stores accumulate. \textbf{Bold}: best cell in each column, the
trained row included. EAGLE-3 cells are our own re-runs on that same machine (KV-cap note as in
Table~\ref{tab:main-echo}). The semantic source is measured against its own absence on GSM8K, the
one of these five for which we ran \ours{}'s lexical-only arm: it moves accepted length by 1.1\%.}
\label{tab:main-8bench}
\resizebox{\textwidth}{!}{%
\small\setlength{\tabcolsep}{3.5pt}
\begin{tabular}{l rr rr rr rr rr r}
\toprule
\textbf{Method} & \multicolumn{2}{c}{\textbf{HumanEval}} & \multicolumn{2}{c}{\textbf{GSM8K}} & \multicolumn{2}{c}{\textbf{MT-Bench}} & \multicolumn{2}{c}{\textbf{WildChat}} & \multicolumn{2}{c}{\textbf{ClassEval}} & \textbf{Avg.} \\
\cmidrule(lr){2-3} \cmidrule(lr){4-5} \cmidrule(lr){6-7} \cmidrule(lr){8-9} \cmidrule(lr){10-11} \cmidrule(lr){12-12}
 & Spd & $\tau$ & Spd & $\tau$ & Spd & $\tau$ & Spd & $\tau$ & Spd & $\tau$ & Spd \\
\midrule
\multicolumn{12}{@{}l}{\textit{Llama-3.1-8B}} \\
AR & 1.00$\times$ & 1.00 & 1.00$\times$ & 1.00 & 1.00$\times$ & 1.00 & 1.00$\times$ & 1.00 & 1.00$\times$ & 1.00 & 1.00$\times$ \\
GOOSE~\citep{goose2026} & 2.10$\times$ & 2.61 & 2.19$\times$ & 2.73 & 1.80$\times$ & 2.23 & 2.10$\times$ & 2.58 & 3.18$\times$ & 4.08 & 2.27$\times$ \\
Token Recycling~\citep{luo2024recycling} & 1.93$\times$ & 2.60 & 2.05$\times$ & 2.76 & 1.73$\times$ & 2.30 & 1.83$\times$ & 2.46 & 2.08$\times$ & 2.83 & 1.92$\times$ \\
SuffixDecoding~\citep{oliaro2024suffixdecoding} & 2.02$\times$ & 2.18 & 1.81$\times$ & 1.96 & 1.34$\times$ & 1.42 & 1.78$\times$ & 1.88 & 3.04$\times$ & 3.38 & 2.00$\times$ \\
\ours{} (Ours) & 2.54$\times$ & 3.34 & 2.24$\times$ & 3.07 & 1.77$\times$ & 2.28 & \textbf{2.24$\times$} & 2.89 & \textbf{3.77$\times$} & 5.08 & 2.51$\times$ \\
\rowcolor{gray!15}
EAGLE-3 (\emph{trained}) & \textbf{3.52$\times$} & \textbf{5.78} & \textbf{3.44$\times$} & \textbf{5.69} & \textbf{3.28$\times$} & \textbf{5.36} & 1.90$\times$ & \textbf{3.09} & 3.00$\times$ & \textbf{5.12} & \textbf{3.03$\times$} \\
\midrule
\multicolumn{12}{@{}l}{\textit{Qwen3-8B}} \\
AR & 1.00$\times$ & 1.00 & 1.00$\times$ & 1.00 & 1.00$\times$ & 1.00 & 1.00$\times$ & 1.00 & 1.00$\times$ & 1.00 & 1.00$\times$ \\
GOOSE~\citep{goose2026} & 2.06$\times$ & 2.54 & 2.05$\times$ & 2.51 & 1.64$\times$ & 2.05 & 1.69$\times$ & 2.23 & 2.55$\times$ & 3.47 & 2.00$\times$ \\
Token Recycling~\citep{luo2024recycling} & 1.85$\times$ & 2.47 & 1.93$\times$ & 2.55 & 1.61$\times$ & 2.15 & 1.55$\times$ & 2.21 & 1.83$\times$ & 2.66 & 1.75$\times$ \\
SuffixDecoding~\citep{oliaro2024suffixdecoding} & 1.92$\times$ & 2.04 & 1.92$\times$ & 2.02 & 1.31$\times$ & 1.37 & 1.44$\times$ & 1.61 & 2.46$\times$ & 2.87 & 1.81$\times$ \\
\ours{} (Ours) & \textbf{2.36$\times$} & 3.09 & 2.33$\times$ & 3.15 & 1.62$\times$ & 2.12 & \textbf{1.71$\times$} & \textbf{2.39} & \textbf{3.04$\times$} & \textbf{4.38} & \textbf{2.21$\times$} \\
\rowcolor{gray!15}
EAGLE-3 (\emph{trained}) & 2.23$\times$ & \textbf{3.55} & \textbf{2.35$\times$} & \textbf{3.69} & \textbf{1.93$\times$} & \textbf{3.09} & 1.40$\times$ & 2.36 & 2.09$\times$ & 3.72 & 2.00$\times$ \\
\bottomrule
\end{tabular}}\end{table*}

\section{Additional Diagnostic Results}
\label{app:prov}\label{sec:mech}\label{app:cases}

\begin{table}[t]
\centering\small
\setlength{\tabcolsep}{6pt}
\begin{tabular}{@{}l rr@{}}
\toprule
 & \textbf{Llama} & \textbf{Qwen3} \\
\midrule
\multicolumn{3}{@{}l}{\textbf{The diagnosis: three questions per position, \%}} \\
Available (verbatim in an earlier trace) & 91.5 & 91.5 \\
Lexically producible (oracle) & 86.8 & 86.0 \\
Available yet not producible (the gap) & 6.81 & 7.52 \\
\midrule
\addlinespace
\multicolumn{3}{@{}l}{\textbf{Why it fails: context still matched, tokens}} \\
Median where the oracle succeeds & 16 & 16 \\
Median at recovered gap positions & 1 & 1 \\
\midrule
\addlinespace
\multicolumn{3}{@{}l}{\textbf{When it fails: gap rate by preceding token}} \\
After a token in no past trace, \% & 29.9 & 29.1 \\
Base rate, \% & 5.50 & 6.27 \\
After a producible token, relative & 0.27$\times$ & 0.32$\times$ \\
\midrule
\addlinespace
\multicolumn{3}{@{}l}{\textbf{Recovery: gap recall by key, \% of gap positions}} \\
Semantic key, online & 80.8 & 83.4 \\
Semantic key, warm pool & 87 & 87 \\
Static token-embedding key & 22.6 & 26.6 \\
\midrule
\addlinespace
\multicolumn{3}{@{}l}{\textbf{What one gap is worth}} \\
Verbatim payload behind it, tokens & 6.8 & 5.5 \\
\quad matched by the neighbor, tokens & 6.4 & 5.3 \\
Blind at lexical-union keys, \% & 4.36 & 4.87 \\
\quad after adding the hidden key, \% & 1.05 & 1.02 \\
Mean uninterrupted run, tokens & 14.7 & 12.9 \\
\quad with the hidden key, tokens & 31.1 & 28.3 \\
Blind-after-blind rate, \% & 15.7 & 20.7 \\
\bottomrule
\end{tabular}
\caption{\textbf{One benchmark in depth, on both model families.} \S\ref{sec:gap}'s API-Bank
quantities, recomputed for Llama-3.1-8B over 10{,}313 generated positions and Qwen3-8B over 8{,}896;
stores are strictly past unless a row says warm. The blocks answer, in order: how often the
continuation is present and how often exact matching can address it (the census itself); how much
context the exact key was still matching where it works and at the gap positions---sixteen tokens
against one, which is why the failure is a cliff and not paraphrase; when gap positions arrive,
which is right after a token no earlier request contained; how much of the gap each key recovers;
and what one recovered position is worth---the verbatim run behind it, how far the draft then gets
before the next blind position, and how often blind positions arrive back to back. The body quotes
the rounded Llama values; every ordering replicates on Qwen3.}
\label{tab:mech}
\end{table}

\begin{table}[t]\centering\small
\setlength{\tabcolsep}{2.6pt}\renewcommand{\arraystretch}{0.95}
\begin{tabular}{@{}l rrr rrr@{}}
\toprule
& \multicolumn{3}{c}{\textit{Llama-3.1-8B}} & \multicolumn{3}{c@{}}{\textit{Qwen3-8B}} \\
\cmidrule(lr){2-4}\cmidrule(l){5-7}
Benchmark & Avail. & Gap & Recall & Avail. & Gap & Recall \\
\midrule
API-Bank & 92 & 6.8 & 81 & 91 & 7.5 & 83 \\
tau-bench retail & 81 & 13.4 & 42 & 84 & 11.7 & 50 \\
tau-bench air & 57 & 18.9 & 29 & 58 & 22.4 & 41 \\
ToolAlpaca & 65 & 24.8 & 47 & 76 & 19.0 & 69 \\
HumanEval & 63 & 27.8 & 42 & 60 & 28.0 & 70 \\
ClassEval & 47 & 25.7 & 45 & 48 & 26.2 & 76 \\
GSM8K & 51 & 31.6 & 55 & 66 & 27.1 & 80 \\
SpecBench-math & 46 & 30.6 & 58 & 61 & 28.0 & 80 \\
MT-Bench & 26 & 21.5 & 9 & 29 & 22.2 & 39 \\
WildChat & 33 & 15.4 & 14 & 36 & 17.4 & 38 \\
\bottomrule
\end{tabular}
\caption{\textbf{The same three questions, across ten benchmarks.} Table~\ref{tab:mech} is one
benchmark in depth; this is how far the diagnosis travels. Percentages of generated positions, stores
strictly past. \emph{Avail.}: the next two tokens occur verbatim in an earlier trace.
\emph{Gap}: available yet not produced by the exact-match oracle (the two predicates differ in horizon,
so the columns do not subtract). \emph{Recall}: gap positions where the semantic key surfaces the
true next token. Wilson
95\% intervals on the Llama API-Bank row: 6.3--7.3\% on Gap ($n{=}10{,}313$), 77.7--83.5\% on Recall
($n{=}702$). The gap is smallest, and the share of it the semantic key recovers highest, where
traffic repeats most; on chat and open-ended text the gap is there but the key reaches almost none
of it. Positions inside one request are not independent draws, so those intervals are
optimistic; resampling whole requests instead (2{,}000 bootstrap draws over the 198) widens them to
5.9--7.8\% and 76.7--84.8\%. SpecBench-math resamples 80 GSM8K instances, so those two rows are not
independent.}
\label{tab:census-grid}
\end{table}

Two requests printed whole from the system's own per-cycle log, reconstructed by an audit script: the API-Bank case behind Figure~\ref{fig:gapproblem}
(Figure~\ref{fig:case-toolspec-apibank}), and a GSM8K control whose values are computed rather than
remembered, so no key of any kind can reach them (Figure~\ref{fig:case-gsm8k}).

\definecolor{csSpine}{HTML}{1D437D}
\definecolor{csTR}{HTML}{0D7B34}
\definecolor{csPLD}{HTML}{0099AD}
\definecolor{csAR}{HTML}{EE442F}
\definecolor{csBar}{HTML}{808080}
\definecolor{csMarkup}{HTML}{3D85C6}
\definecolor{csBand}{HTML}{DDE3EC}
\newcommand{\csLabel}[1]{{\normalfont\sffamily\scriptsize\bfseries\color{csSpine}\MakeUppercase{#1}}}
\newcommand{\csElide}[1]{{\normalfont\sffamily\tiny\itshape\color{black!45}#1}}

\begin{figure*}[!t]
\noindent\fcolorbox{black!20}{white}{%
\begin{minipage}{\dimexpr\textwidth-2\fboxsep-2\fboxrule}
\setlength{\parindent}{0pt}\ttfamily\scriptsize\raggedright\setlength{\baselineskip}{8.4pt}\color{black!72}
\csLabel{SYSTEM PROMPT}\\[1pt]
\textcolor{black!25}{\rule{\linewidth}{0.4pt}}\\[2pt]
\mbox{You~are~a~helpful~multi-turn~dialogue~assistant~capable~of~leveraging~tool~calls~to~solve~user~tasks~and}\\
\mbox{~~~~~provide~structured~chat~responses.}\\
\mbox{}\\
\mbox{\textcolor{csMarkup}{**Available~Tools**}}\\
\mbox{In~your~response,~you~can~use~the~following~tools:}\\
\mbox{\csElide{[ 13 lines / 654 characters of tool definitions elided here ]}}\\
\mbox{}\\
\mbox{\textcolor{csMarkup}{**Important~Notes**}}\\
\mbox{1.~Provide~at~least~one~of~`\textcolor{csMarkup}{<tool\_call>}`.}\\
\mbox{2.~You~can~invoke~multiple~tool~calls~simultaneously~in~the~`\textcolor{csMarkup}{<tool\_call>}`~fields.~Each~tool~call~should~}\\
\mbox{~~~~be~a~JSON~object~with~a~"name"~field~and~an~"parameters"~field~containing~a~dictionary~of~parame~~[<}\\
\mbox{~~~~84~more~characters~on~this~line>]}\\
\mbox{3.~Refer~to~the~previous~dialogue~records~in~the~history,~including~the~user's~queries,~previous~`\textcolor{csMarkup}{<tool\_}}\\
\mbox{~~~~\textcolor{csMarkup}{call>}`,~`\textcolor{csMarkup}{<response>}`,~and~any~tool~feedback~noted~as~`\textcolor{csMarkup}{<obs>}`~(if~exists).}\\
\vspace{5pt}
\csLabel{USER TURN / CONVERSATION SO FAR}\\[1pt]
\textcolor{black!25}{\rule{\linewidth}{0.4pt}}\\[2pt]
\mbox{\textcolor{csMarkup}{**Dialogue~Records~History**}}\\
\mbox{\textcolor{csMarkup}{<user>}Book~a~meeting~for~me,~including~all~employees~in~the~Alibaba~who~are~not~traveling.~Today~is~2023}\\
\mbox{~~~~.6.8,~the~meeting~is~from~14:00~to~15:00.\textcolor{csMarkup}{</user>}}\\
\mbox{\textcolor{csMarkup}{<tool\_call>}}\\
\mbox{\{"name":~"OrganizationMembers",~"parameters":~\{"organization":~"Alibaba"\}\}}\\
\mbox{\textcolor{csMarkup}{</tool\_call>}}\\
\mbox{\textcolor{csMarkup}{<obs>}~\{'api\_name':~'OrganizationMembers',~'input':~\{'organization':~'Alibaba'\},~'output':~\{'members':~['}\\
\mbox{~~~~John',~'Mary',~'Peter']\},~'exception':~None\}~\textcolor{csMarkup}{</obs>}}\\
\mbox{\textcolor{csMarkup}{<tool\_call>}}\\
\mbox{\{"name":~"TravelStatus",~"parameters":~\{"member\_name":~"John"\}\}}\\
\mbox{\textcolor{csMarkup}{</tool\_call>}}\\
\mbox{\textcolor{csMarkup}{<obs>}~\{'api\_name':~'TravelStatus',~'input':~\{'member\_name':~'John'\},~'output':~'Traveling',~'exception':}\\
\mbox{~~~~~None\}~\textcolor{csMarkup}{</obs>}}\\
\mbox{}\\
\mbox{\textcolor{csMarkup}{<user>}~Based~on~our~conversation~above,~please~only~make~one~tool~call~to~solve~my~need.\textcolor{csMarkup}{</user>}}\\
\mbox{}\\
\vspace{5pt}
\colorbox{csBand!55}{\begin{minipage}{\dimexpr\linewidth-2\fboxsep}\ttfamily\scriptsize\raggedright\setlength{\baselineskip}{8.4pt}
\csLabel{MODEL OUTPUT}\\[1pt]
\textcolor{black!25}{\rule{\linewidth}{0.4pt}}\\[2pt]
\mbox{\textcolor{csSpine}{<tool\_call>}}\\
\mbox{\textcolor{csSpine}{\{"name":~"}\textcolor{csBar}{|}\textcolor{csSpine}{OrganizationMembers",~"parameters":~\{"organization":~"Alibaba"\}\}}}\\
\mbox{\textcolor{csSpine}{</tool\_call>}}\\
\mbox{\textcolor{csSpine}{<}\textcolor{csBar}{|}\textcolor{csSpine}{tool\_call>}}\\
\mbox{\textcolor{csSpine}{\{"name":~"}\textcolor{csBar}{|}\textcolor{csSpine}{TravelStatus",~"parameters":~\{"member\_name":~"John"\}\}}}\\
\mbox{\textcolor{csSpine}{</tool\_call>}}\\
\mbox{\textcolor{csSpine}{<}\textcolor{csBar}{|}\textcolor{csSpine}{tool\_call>}}\\
\mbox{\textcolor{csSpine}{\{"name":~"TravelStatus",~"parameters":~\{"member}\textcolor{csBar}{|}\textcolor{csSpine}{\_name":~"}\textcolor{csBar}{|}\textcolor{csAR}{Mary}\textcolor{csBar}{|}\textcolor{csTR}{"\}\}}}\\
\mbox{\textcolor{csTR}{</tool\_call}\textcolor{csBar}{|}\textcolor{csSpine}{>}}\\
\mbox{\textcolor{csSpine}{<tool\_call>}}\\
\mbox{\textcolor{csBar}{|}\textcolor{csSpine}{\{"name":~"TravelStatus",~"parameters":~\{"member\_name":~"}\textcolor{csBar}{|}\textcolor{csTR}{Peter"\}\}}}\\
\mbox{\textcolor{csTR}{</tool\_call}\textcolor{csBar}{|}\textcolor{csSpine}{>}}\\
\mbox{\textcolor{csSpine}{<tool\_call}\textcolor{csBar}{|}\textcolor{csSpine}{>}}\\
\mbox{\textcolor{csSpine}{\{"name":~"Booking}\textcolor{csBar}{|}\textcolor{csTR}{Meeting",~"}\textcolor{csBar}{|}\textcolor{csSpine}{parameters":~\{"}\textcolor{csBar}{|}\textcolor{csTR}{date":~"}\textcolor{csBar}{|}\textcolor{csSpine}{2023.}\textcolor{csBar}{|}\textcolor{csSpine}{6.8}\textcolor{csBar}{|}\textcolor{csTR}{",~"start\_time}\textcolor{csBar}{|}\textcolor{csSpine}{":~"14:00}\textcolor{csBar}{|}\textcolor{csSpine}{",~"end\_time":}}\\
\mbox{~~~~\textcolor{csSpine}{~"}\textcolor{csBar}{|}\textcolor{csAR}{15}\textcolor{csBar}{|}\textcolor{csSpine}{:00",~"}\textcolor{csBar}{|}\textcolor{csTR}{members":~["John",}\textcolor{csBar}{|}\textcolor{csAR}{~"}\textcolor{csBar}{|}\textcolor{csSpine}{Mary",~"}\textcolor{csBar}{|}\textcolor{csAR}{Peter}\textcolor{csBar}{|}\textcolor{csAR}{"]}\textcolor{csBar}{|}\textcolor{csTR}{\}\}}}\\
\mbox{\textcolor{csTR}{</tool\_call>}}\\
\end{minipage}}
\end{minipage}}
\caption{\textbf{API-Bank, one request end to end.} The running example of Figure~\ref{fig:gapproblem}: the value the lexical key goes blind on is a colleague's name that the conversation already supplied. Pool warmed on the benchmark's first 20 requests; prompt 550 tokens, 160 generated in 30 verification passes ($\tau{=}5.33$ for this request alone). Tokens committed, by the source that supplied them, a pass whose draft is rejected outright committing only the verifier's own token: \textcolor{csSpine}{merged tree} 126, \textcolor{csTR}{Token Recycling} 29, \textcolor{csAR}{verifier only} 5. Vertical bars mark span boundaries and are \emph{not} in the model's output; lines over 104 characters are wrapped by us. Truncated at this trace's 160-token cap.}
\label{fig:case-toolspec-apibank}
\end{figure*}

\begin{figure*}[!t]
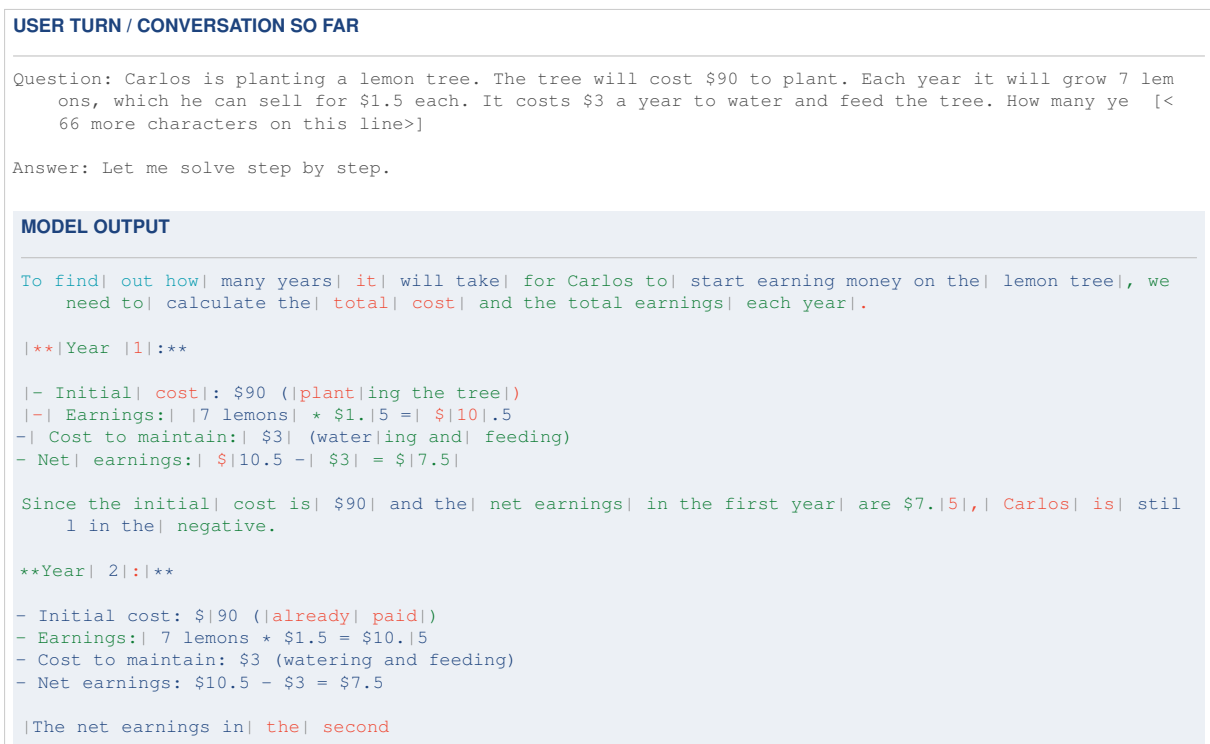

\noindent\fcolorbox{black!20}{white}{%
\begin{minipage}{\dimexpr\textwidth-2\fboxsep-2\fboxrule}
\setlength{\parindent}{0pt}\ttfamily\scriptsize\raggedright\setlength{\baselineskip}{8.4pt}\color{black!72}
\csLabel{USER TURN / CONVERSATION SO FAR}\\[1pt]
\textcolor{black!25}{\rule{\linewidth}{0.4pt}}\\[2pt]
\mbox{Question:~Carlos~is~planting~a~lemon~tree.~The~tree~will~cost~\$90~to~plant.~Each~year~it~will~grow~7~lem}\\
\mbox{~~~~ons,~which~he~can~sell~for~\$1.5~each.~It~costs~\$3~a~year~to~water~and~feed~the~tree.~How~many~ye~~[<}\\
\mbox{~~~~66~more~characters~on~this~line>]}\\
\mbox{}\\
\mbox{Answer:~Let~me~solve~step~by~step.}\\
\mbox{}\\
\vspace{5pt}
\colorbox{csBand!55}{\begin{minipage}{\dimexpr\linewidth-2\fboxsep}\ttfamily\scriptsize\raggedright\setlength{\baselineskip}{8.4pt}
\csLabel{MODEL OUTPUT}\\[1pt]
\textcolor{black!25}{\rule{\linewidth}{0.4pt}}\\[2pt]
\mbox{\textcolor{csPLD}{To~find}\textcolor{csBar}{|}\textcolor{csPLD}{~out~how}\textcolor{csBar}{|}\textcolor{csSpine}{~many~years}\textcolor{csBar}{|}\textcolor{csAR}{~it}\textcolor{csBar}{|}\textcolor{csSpine}{~will~take}\textcolor{csBar}{|}\textcolor{csTR}{~for~Carlos~to}\textcolor{csBar}{|}\textcolor{csSpine}{~start~earning~money~on~the}\textcolor{csBar}{|}\textcolor{csSpine}{~lemon~tree}\textcolor{csBar}{|}\textcolor{csTR}{,~we~}}\\
\mbox{~~~~\textcolor{csTR}{need~to}\textcolor{csBar}{|}\textcolor{csSpine}{~calculate~the}\textcolor{csBar}{|}\textcolor{csAR}{~total}\textcolor{csBar}{|}\textcolor{csAR}{~cost}\textcolor{csBar}{|}\textcolor{csTR}{~and~the~total~earnings}\textcolor{csBar}{|}\textcolor{csTR}{~each~year}\textcolor{csBar}{|}\textcolor{csAR}{.}}\\
\mbox{}\\
\mbox{\textcolor{csBar}{|}\textcolor{csAR}{**}\textcolor{csBar}{|}\textcolor{csTR}{Year~}\textcolor{csBar}{|}\textcolor{csAR}{1}\textcolor{csBar}{|}\textcolor{csSpine}{:**}}\\
\mbox{}\\
\mbox{\textcolor{csBar}{|}\textcolor{csTR}{-~Initial}\textcolor{csBar}{|}\textcolor{csAR}{~cost}\textcolor{csBar}{|}\textcolor{csSpine}{:~\$90~(}\textcolor{csBar}{|}\textcolor{csAR}{plant}\textcolor{csBar}{|}\textcolor{csTR}{ing~the~tree}\textcolor{csBar}{|}\textcolor{csAR}{)}}\\
\mbox{\textcolor{csBar}{|}\textcolor{csAR}{-}\textcolor{csBar}{|}\textcolor{csTR}{~Earnings:}\textcolor{csBar}{|}\textcolor{csAR}{~}\textcolor{csBar}{|}\textcolor{csSpine}{7~lemons}\textcolor{csBar}{|}\textcolor{csTR}{~*~\$1.}\textcolor{csBar}{|}\textcolor{csSpine}{5~=}\textcolor{csBar}{|}\textcolor{csAR}{~\$}\textcolor{csBar}{|}\textcolor{csAR}{10}\textcolor{csBar}{|}\textcolor{csSpine}{.5}}\\
\mbox{\textcolor{csSpine}{-}\textcolor{csBar}{|}\textcolor{csTR}{~Cost~to~maintain:}\textcolor{csBar}{|}\textcolor{csSpine}{~\$3}\textcolor{csBar}{|}\textcolor{csSpine}{~(water}\textcolor{csBar}{|}\textcolor{csTR}{ing~and}\textcolor{csBar}{|}\textcolor{csTR}{~feeding)}}\\
\mbox{\textcolor{csTR}{-~Net}\textcolor{csBar}{|}\textcolor{csTR}{~earnings:}\textcolor{csBar}{|}\textcolor{csAR}{~\$}\textcolor{csBar}{|}\textcolor{csSpine}{10.5~-}\textcolor{csBar}{|}\textcolor{csTR}{~\$3}\textcolor{csBar}{|}\textcolor{csTR}{~=~\$}\textcolor{csBar}{|}\textcolor{csTR}{7.5}\textcolor{csBar}{|}}\\
\mbox{}\\
\mbox{\textcolor{csTR}{Since~the~initial}\textcolor{csBar}{|}\textcolor{csTR}{~cost~is}\textcolor{csBar}{|}\textcolor{csSpine}{~\$90}\textcolor{csBar}{|}\textcolor{csSpine}{~and~the}\textcolor{csBar}{|}\textcolor{csTR}{~net~earnings}\textcolor{csBar}{|}\textcolor{csTR}{~in~the~first~year}\textcolor{csBar}{|}\textcolor{csTR}{~are~\$7.}\textcolor{csBar}{|}\textcolor{csAR}{5}\textcolor{csBar}{|}\textcolor{csAR}{,}\textcolor{csBar}{|}\textcolor{csAR}{~Carlos}\textcolor{csBar}{|}\textcolor{csAR}{~is}\textcolor{csBar}{|}\textcolor{csSpine}{~stil}}\\
\mbox{~~~~\textcolor{csSpine}{l~in~the}\textcolor{csBar}{|}\textcolor{csTR}{~negative.}}\\
\mbox{}\\
\mbox{\textcolor{csTR}{**Year}\textcolor{csBar}{|}\textcolor{csSpine}{~2}\textcolor{csBar}{|}\textcolor{csAR}{:}\textcolor{csBar}{|}\textcolor{csSpine}{**}}\\
\mbox{}\\
\mbox{\textcolor{csSpine}{-~Initial~cost:~\$}\textcolor{csBar}{|}\textcolor{csSpine}{90~(}\textcolor{csBar}{|}\textcolor{csAR}{already}\textcolor{csBar}{|}\textcolor{csAR}{~paid}\textcolor{csBar}{|}\textcolor{csTR}{)}}\\
\mbox{\textcolor{csTR}{-~Earnings:}\textcolor{csBar}{|}\textcolor{csSpine}{~7~lemons~*~\$1.5~=~\$10.}\textcolor{csBar}{|}\textcolor{csSpine}{5}}\\
\mbox{\textcolor{csSpine}{-~Cost~to~maintain:~\$3~(watering~and~feeding)}}\\
\mbox{\textcolor{csSpine}{-~Net~earnings:~\$10.5~-~\$3~=~\$7.5}}\\
\mbox{}\\
\mbox{\textcolor{csBar}{|}\textcolor{csSpine}{The~net~earnings~in}\textcolor{csBar}{|}\textcolor{csAR}{~the}\textcolor{csBar}{|}\textcolor{csAR}{~second}}\\
\end{minipage}}
\end{minipage}}
\caption{\textbf{GSM8K, one request end to end.} The control: the numbers the model emits are computed from this problem and appear nowhere in the pool, so there is nothing for any copy method to find. Pool warmed on the benchmark's first 12 requests; prompt 112 tokens, 200 generated in 70 verification passes ($\tau{=}2.86$ for this request alone). Tokens committed, by the source that supplied them, a pass whose draft is rejected outright committing only the verifier's own token: \textcolor{csSpine}{merged tree} 102, \textcolor{csTR}{Token Recycling} 71, \textcolor{csPLD}{prompt lookup} 4, \textcolor{csAR}{verifier only} 22, \textcolor{csAR}{unaccounted tail} 1. Vertical bars mark span boundaries and are \emph{not} in the model's output; lines over 104 characters are wrapped by us. Truncated at this trace's 200-token cap.}
\label{fig:case-gsm8k}
\end{figure*}

\end{document}